\pdfoutput=1
\documentclass[journal]{IEEEtran}
\usepackage{times}
\usepackage{soul}
\usepackage{url}
\usepackage{hyperref}
\usepackage[utf8]{inputenc}
\usepackage[small]{caption}
\usepackage{amsfonts,amssymb}
\usepackage{graphicx}
\usepackage{amsmath}
\usepackage{booktabs}
\usepackage{algorithm}
\usepackage{algorithmic}
\usepackage{multirow}
\usepackage{xspace}
\usepackage{subfigure}
\usepackage{footnote}% Use the postscript times font!
\usepackage{color}
\usepackage{setspace}

\let\oldhat\hat
\renewcommand{\vec}[1]{\mathbf{#1}}
\renewcommand{\hat}[1]{\oldhat{\mathbf{#1}}}
\renewcommand{\matrix}[1]{\mathbf{#1}}

\usepackage{epstopdf}
\usepackage{threeparttable}
\usepackage{dcolumn}
\newcolumntype{d}[1]{D{.}{.}{#1}}% or D{.}{,}{#1} or D{.}{\cdot}{#1}

\newcommand{\eg}{\emph{e.g.,}\xspace}
\newcommand{\rf}{\emph{cf}\xspace}

\newcommand{\ie}{\emph{i.e.,}\xspace}
\newcommand{\etc}{\emph{etc.}\xspace}
\newcommand{\etal}{\emph{et al.}\xspace}

\newcommand{\eat}[1]{}
\newcommand{\paratitle}[1]{\vspace{1ex}\noindent \textbf{#1}}
\begin{document}

\title{Position-Aware Self-Attention based  Neural Sequence Labeling}

\author{Wei Wei,
Zanbo Wang,
Xianling Mao,
Guangyou Zhou,
Pan Zhou,
and Sheng Jiang
%and Kun He,~\IEEEmembership{Senior Member,~IEEE}
%
%
% \thanks{Manuscript received XXX X, XXXX; revised XXX X, XXXXa.}
\thanks{W. Wei (corresponding author), Z.-B.~Wang and S. Jiang (corresponding author) are with the School of Computer Science and Technology, Huazhong University of Science and Technology (HUST), Wuhan, Hubei, 430074, China. E-mail: weiw@hust.edu.cn (or weiwei8329@gmail.com)}.
\thanks{X-L. Mao is with School of Computer, Beijing Institute of Technology, Beijing 100081, China. E-mail: {maoxl}@bit.edu.cn.}
\thanks{G.-Y. Zhou is with  School of Computer, Central China Normal University, Wuhan, Hubei, 430000 China. E-mail: {gyzhou}@mail.ccnu.edu.cn.}
\thanks{P. Zhou are with School of Electronic Information and Communications, Huazhong University of Science and Technology (HUST), Wuhan, Hubei, 430074, P.R. China.}

%
%\thanks{H. Li is with Noah's Ark Lab, Huawei Technologies 5F, Core Building Two, HK Science Park, Shatin, New Territories, Hong Kong. E-mail: hangli.hl@huawei.com.}
%\thanks{An earlier version of this article was presented at SIGKDD 2011, titled Semi-Supervised Ranking on Very Large Graphs with Rich Metadata.}
}

\maketitle

% make the title area
% \maketitle

% As a general rule, do not put math, special symbols or citations
% in the abstract or keywords.
\begin{abstract}
  \emph{Sequence labeling} is a fundamental task in natural language processing and has been widely studied. Recently, RNN-based sequence labeling models have increasingly gained attentions. Despite superior performance achieved by learning the long short-term (\ie \emph{successive}) dependencies,
  %long term token dependencies,
  the way of sequentially processing inputs might limit the ability to capture the non-continuous relations over tokens within a sentence.
  To tackle the problem, we focus on how to effectively model \emph{successive} and \emph{discrete} dependencies of each token for enhancing the sequence labeling performance.
  %which is rarely addressed in previous works.
  %
  Specifically, we propose an innovative attention-based model (called \textsf{position-aware self-attention}, \ie \textsf{PSA}) as well as a well-designed
  self-attentional context fusion layer within a neural network architecture, to explore the positional information of an input sequence for capturing the latent relations among tokens.
  Extensive experiments on three classical tasks in \emph{sequence labeling} domain, \ie~\emph{part-of-speech} (\textbf{POS}) \emph{tagging}, \emph{named entity recognition} (\textbf{NER}) and \emph{phrase chunking}, demonstrate our proposed model outperforms the state-of-the-arts without any external knowledge, in
  terms of various metrics.
  % the significant improvements of  over  methods without any external knowledge.

\end{abstract}

% Note that keywords are not normally used for peerreview papers.

\section{Introduction}
\emph{Sequence labeling}, named \textsf{SL}, is one of \emph{pattern recognition} task
in the filed of natural language processing (NLP) and machine learning (ML),
which aims to assign a categorical label to each element of a sequence of observed values,
such as \emph{part-of-speech} (POS) \emph{tagging}~\cite{Ma2016End}, \emph{chunking}~\cite{Liu2017Empower}
and \emph{named entity recognition} (NER)~\cite{Lample2016Neural} and \etc
It plays a pivotal role in
%the domain of
\emph{natural language understanding} (NLU)
and significantly beneficial for a variety of \emph{downstream} applications, \eg \emph{syntactic parsing},
\emph{relation extraction} and \emph{entity coreference resolution} and \etc

Conventional sequence labeling approaches are usually
on the basis of classical machine learning technologies,
such as \emph{Hidden Markov Models} (HMM)~\cite{baum1966statistical} and \emph{Conditional Random Fields} (CRF)~\cite{lafferty2001conditional},
which heavily rely on
%relying on many
hand-crafted features (\eg with/without capitalized word)
%whether a word is capitalized) 
or language-specific resources (\eg gazetteers),
making it difficult to apply them to new language-related tasks or domains.
With advances in deep learning,
many research efforts have been dedicated to
enhancing \textsf{SL}~\cite{Ma2016End} by
automatically extracting features via
%utilizing
different types of neural networks (NNs),
%to automatically extract features,
where various characteristics of word information are encoded in distributed representations for inputs~\cite{Collobert2011Natural}
and the sentence-level context representations are learned when end-to-end training.

\begin{figure}[!t]
	\centering
	\includegraphics[width=0.5\textwidth]{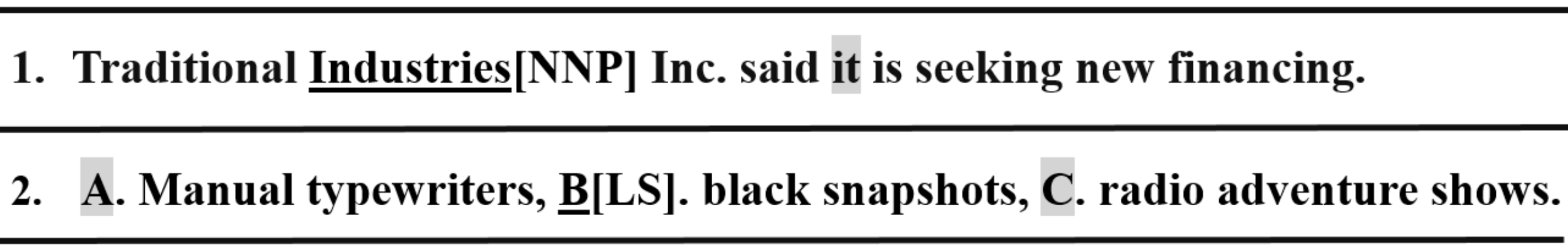}
	\caption{Example: The impacts of Discrete Context Dependencies.}
	\label{fig:motivation}
	\vspace{-2ex}
\end{figure}

%\begin{figure}[!t]
%\centering
%\includegraphics[width=0.3\textwidth]{figures/motivation.pdf}
%\vspace{-3pt}\caption{Example: The impact of discrete context dependencies.}
%\label{fig:motivation}
%\vspace{0pt}
%\end{figure}

Recently, Recurrent Neural Network (RNN) together with its variants, \eg long short-term memory (LSTM) or gated recurrent unit (GRU), have shown great success in modeling sequential data~\cite{yousfi2017contribution}.
Therefore, many researches have devoted to research on RNN based architectures for \textsf{SL}, such as BiLSTM-CNN~\cite{chiu2016named}, LSTM-CRF~\cite{lee2018effective,kwon2019effective}, LSTM-CNN-CRF~\cite{Ma2016End} and \etc
Despite superior performance achieved, these models have limitations under the fact that RNNs recursively compose each word with its previous hidden state encoded with the entire history information,
but the latent independent relations between each pair of words are not well managed. The sequential way to process the inputs only focuses on modeling the long-range \emph{successive} context dependencies, while neglecting the \emph{discrete} context patterns\footnote{In the case study part of Section \ref{sec:exp}, we present multiple error cases to help explain and prove that the RNN models may cause insufficient modelling of discrete context dependency.}.

Discrete context dependency
%is an important feature for sequence labeling tasks.
%
plays a significant role in sequence labeling tasks.
Generally, for a given word, its label not only depends on its own semantic information and neighbor contexts,
but may also rely on the separate word information within the same sequences,
which would significantly affect the accuracy of labeling.
Without loss of generality, we take the part-of-speech (POS) tagging task as example, as shown in Figure~\ref{fig:motivation},
the part-of-speech tag of word ``Industries" in \emph{Sentence-1} primarily depends on word ``it", and thus should
label with NNP, which refers to \emph{singular proper noun}. However, if such discrete context dependency is not
well modeled, ``Industries" may tend to be labeled with \emph{plural proper noun} (NNPS) mistakenly,
since a word ending with ``-s'' and more so ``-es'' is more likely labeled with NNPS.
%But if this kind of discrete context dependency is not well modeled, ``Industries" tends to be labeled as \emph{plural proper noun} (NNPS) mistakenly because `s' and more so `es' is a common way to end a plural noun.
%
Similar to \emph{Sentence-2}, assigning the part-of-speech tag \emph{list item marker} (LS) to word ``B" should take account of word ``A" and ``C", where these three constitute a list.
Therefore, it is essential to selectively choose the contexts that have strong impacts on the tag of the given word.

Many works demonstrate that self-attention is capable of effectively improving the performance 
of several NLP tasks such as machine translation, reading comprehension and semantic role labeling.
%Recently, self-attention has been successfully applied to many NLP tasks for performance improvement~\cite{Tan2018Deep,Cheng2016Long,Vaswani2017Attention,lin2017structured},
%
%such as  semantic role labeling~\cite{Tan2018Deep}, reading comprehension~\cite{Cheng2016Long} and machine translation~\cite{Vaswani2017Attention},
% Self-attention is actually an attention mechanism of computing a categorical distribution of the input sequence
% via capturing dependencies over tokens.
%to compute a categorical distribution of the input sequence for capturing dependencies over tokens.
%
This inspires us to introduce \emph{self-attention} to explicitly model \emph{position}-aware contexts of a given sequence.
%
%the effectiveness of which has been proven by Vaswani \etal~\cite{Vaswani2017Attention} that
%Vaswani \etal have proven that encoding \emph{absolute} positions of the input sequence with attentions is

Although encoding \emph{absolute} positions of the input sequence with attentions~\cite{Vaswani2017Attention} has been proven the effectiveness,
compared with injecting \emph{absolute} position embedding into the initial representations, it is more intuitive to 
incorporate the positional information in a \emph{relative} manner.
%that encoding \emph{absolute} positions of the input sequence with attentions is benefit for sequence labeling.
%
Recently, Shaw \etal~\cite{Shaw2018Self} present an alternative approach 
to take the account of the \emph{relative} distance between sequence elements
for representation.
%
%In addition, self-attention is a special case of attention mechanism that compute a categorical distribution of input sequence for capturing dependencies over the tokens.
%
%Self-attention has been successfully applied to many tasks for performance improvement in recent years, such as, reading comprehension~\cite{Cheng2016Long}, textual entailment~\cite{lin2017structured}, machine translation~\cite{Vaswani2017Attention} and semantic role labeling~\cite{Tan2018Deep} ~\emph{etc.}.
%In contrast to recurrent and convolutional neural networks, a disadvantage of most attention mechanisms is that it does not explicitly model position information in its structure. Vaswani et al.~\cite{Vaswani2017Attention} first point out the problem and address it by applying absolute position encoding to the input sequence before being processed by attention. Shaw et al.~\cite{Shaw2018Self} subsequently present an alternative approach that extends the self-attention mechanism to efficiently consider representations of the relative distances between sequence elements.
%
Nevertheless, their approaches only consider the relative position information
independent of the sequence tokens while neglecting the interaction with the input representations.
Hence, how to effectively exploit the position information with attentions for better modeling the context dependency is still an open problem.
%these approaches considers the absolute or relative position information independent of the sequence tokens and neglects its interactions with the input representations. How to model position information within attention mechanism by a more flexible manner and favor to the modeling of context dependency is still an open problem.

In this paper, we propose a novel RNN neural architecture for sequence labeling tasks,
which employs self-attention to implicitly encode position information 
to provide complementary context information on the basis of Bi-LSTM.
%rather than directly utilizing the relative distance of each pair of tokens in the process of calculating attention scores,
%it is benefit for inducing the latent relations among such tokens. 
%
Additionally, we further propose an extension of \emph{standard} additive self-attention mechanism (named \textsf{position-aware self-attention}, \textsf{PSA})
to model the discrete context dependencies of the input sequence.
Differ from previous works,
\textsf{PSA} maintains a \emph{variable-length} memory to explore position information in a more flexible manner
for tackling the above mentioned problem.
That is,
it jointly exploits three different positional bias, \ie \emph{self-disabled mask} bias, \emph{distance-aware Gaussian} bias and \emph{token-specific position} bias,
to induce the latent independent relations among tokens,
which can effectively model the discrete context dependencies of given sequence.
Additionally, we also develop a well-designed self-attentional context fusion layer with feature-wise gating mechanism to dynamically select useful information about discrete context dependency and also address the \emph{self-disabled mask} bias problem. Specifically, it
learns a parameter $\lambda$ to adaptively combine the input and the output of the \emph{position}-aware self-attention 
and then generate the context-aware representations of each token.
% In addition, due to \emph{self-disabled mask} bias problem, correspondingly a self-attentional context fusion layer is developed, which learns a parameter $\lambda$ to combine the input and output of the \emph{position}-aware self-attention via fusion gate and generate the context-aware representations of each token.
%
The extensive experiments conducted on four classical benchmark datasets within the domain of \emph{sequence labeling}, 
\ie the CoNLL 2003 NER, the WSJ portion of the Penn Treebank POS tagging, the CoNLL 2000 chunking and the OntoNotes 5.0 English NER,
demonstrate that our proposed model achieves a significant improvement over the state-of-the-arts.
%
%In experiments, we compare our model with various existing top-performance methods on three classical benchmark datasets in the domain of \emph{sequence labeling}, \ie the CoNLL 2003 NER, the WSJ portion of the Penn Treebank POS tagging, as well as the CoNLL 2000 chunking. Experimental results show that our proposed model achieves a significant improvement over the state-of-the-arts.
%
The main contributions of this work are as follows.
%To sum up, our contributions are as follows:
\begin{itemize}
\item We identify the problem of modeling discrete context dependencies in sequence labeling tasks.
\item We propose a novel \emph{position}-aware self-attention to incorporate three different positional factors
for exploring the \emph{relative} position information among tokens;
and also develop a well-designed self-attentional context fusion with feature-wise gating mechanism
to provide complementary context information on the basis of Bi-LSTM for better modeling 
the discrete context dependencies over tokens.
%by incorporating three different positional factors to explore the \emph{relative} position information among tokens and develop a self-attentional context fusion layer to provide complementary context information on the basis of Bi-LSTM.
%\item A new neural architecture for sequence labeling tasks is introduced to better model discrete context dependencies and extract context features of tokens.
\item Extensive experiments on \emph{part-of-speech} (\textbf{POS}) \emph{tagging}, \emph{named entity recognition} (\textbf{NER}) and \emph{phrase chunking} tasks verify the effectiveness of our proposed model.
\end{itemize}

\paratitle{Roadmap}. The remaining of the paper is organized as follows.
In Section \ref{sec:related}, we review the related work,
and in Section \ref{sec:prelimi} we presents a background on sequence labeling tasks, as well as a \emph{Bi-LSTM-CRF} baseline model,
followed with the proposed \emph{position}-aware self-attention mechanism and  self-attentional context fusion layer in Section \ref{sec:method}.
%
%Section \ref{sec:method} first illustrates the novel \emph{position}-aware self-attention mechanism, then presents the proposed self-attentional context fusion layer.
%
Section \ref{sec:exp} presents the quantitative results on benchmark datasets, also includes an in-depth analysis, case study and
wraps up discussion over the obtained results.
Finally, Section \ref{sec:conclusion} concludes the paper.
 
%presents some quantitative results on benchmark datasets to demonstrate the effectiveness of our proposed model. 
%It further presents an in-depth analysis through ablation study and case study and wraps up discussion over the obtained results. Section \ref{sec:related} gives an overview of the related work and Section \ref{sec:conclusion} concludes the paper.

\section{Related Work}
\label{sec:related}

There exist three threads of related work regarding our proposed sequence labeling problem,
%the topics in this paper,
namely, \emph{sequence labeling}, \emph{self-attention}
and \emph{position based attention}.

\subsection{Sequence Labeling}
Sequence labeling is a category of fundamental tasks in natural language processing (NLP),
\eg POS tagging, phrase chunking, named entity recognition (NER) and \etc
Most of conventional high performance sequence labeling approaches are based on classical statistical machine learning models,
such as HMM~\cite{rabiner1989tutorial},
CRFs~\cite{lafferty2001conditional,Mccallum2003Early}, Support Vector Machine (SVM)~\cite{kudoh2000use},
Perceptron~\cite{collins2002discriminative}, and \etc,
where the well-designed features are required for training.
%which require well-designed features extracted from the given data for training.
%
%to represent data and machine learning algorithms are then applied to learn model. Some of the better known algorithms include Hidden Markov Models (HMM)~\cite{rabiner1989tutorial}, Conditional Random Fields (CRF)~\cite{lafferty2001conditional,Mccallum2003Early}, Support Vector Machine (SVM)~\cite{kudoh2000use} and Perceptron~\cite{collins2002discriminative}.
%\eg Hidden Markov Models (HMM)~\cite{rabiner1989tutorial}, Conditional Random Fields (CRF)~\cite{lafferty2001conditional,Mccallum2003Early} and %Perceptron~\cite{collins2002discriminative}.
%~\cite{kudoh2000use} first applies the Support Vector Machine (SVM) model to phrase chunking task, and achieves the best results on CoNLL2000 %dataset at that time. Several
%

Although the great success has been achieved by the traditional supervised
learning based methods,
%
% these methods still fail to achieve good performances due to several facts,
% such as heavily relying on handcrafted features/external domain-specific knowledge,
% or poor generalization ability on the new datasets.
these approaches require a lot of engineering skill 
and domain expertise to design handcrafted features.

With the rise of deep learning,
many research efforts have been conducted on neural network based approaches
to automatically learning the feature representation for \textsf{SL} tasks.
The pioneering work is firstly proposed by Collobert \etal~\cite{Collobert2011Natural}
to extract context-aware features using a simple feed-forward neural network with a fixed-size window,
and generate the final labeled sequence through a CRF layer,
%, who use a simple feed-forward neural network with a fixed-size window for context-aware feature extraction and generate the output label sequence with a CRF layer.
which yields good performance in POS tagging, chunking, NER and \etc
%
%The model achieves good performance on the tasks of part-of-speech tagging, chunking and named entity recognition, with little task-specific engineering done.
%
However, such window-based methods essentially follow  a hypothesis, according to which the tags of an input word mainly depend on its neighboring words, while neglecting the global long-range contexts.
%
%the window approach assumes the tag of a word depends mainly on its neighboring words, which neglects the long-range global context information.

Hence, several variants of bidirectional recurrent neural networks, \eg Bidirectional Long Short-Term Memory (Bi-LSTM) and Bidirectional Gated Recurrent Unit (Bi-GRU), are proposed to encode long-range dependency features for the representation learning of each word, and thus achieve excellent performances.
%
%Recently, the variants of bidirectional recurrent neural networks such as bidirectional long-short term memory (Bi-LSTM) and bidirectional gated recurrent unit (Bi-GRU), have shown great success for modeling long-range dependencies and learning global context-aware representations for words.
%While Hammerton~\cite{hammerton2003named} has studied utilizing LSTMs for NER tasks in the past, the lack of computing power limits the effectiveness of their model.
%With recent advances in deep learning, much research effort has been dedicated to using Bi-RNN for encoding context features in their sequence labeling models and achieve excellent performance.
%
Huang \etal~\cite{Huang2015Bidirectional} initially employ a Bi-LSTM model to encode contextual representations of each word and then
adopt a CRF model to jointly decode.
%
% initially adopt Bi-LSTM to generate contextual representations of every word and use CRF model for jointly label decoding.
%
Subsequently, the proposed Bi-LSTM-CRF architecture is widely used for various sequence labeling tasks.
%The proposed Bi-LSTM-CRF model is almost the most widely used architecture today and has become baseline model for many subsequent work.
Lample \etal~\cite{Lample2016Neural} and Ma \etal~\cite{Ma2016End}
both extend such model with an additional LSTM/CNN layer to encode character-level representations.
%extend the Bi-LSTM-CRF model by adding a LSTM/CNN layer to obtain character-level representations, respectively.
Liu \etal~\cite{Liu2017Empower} conduct a multi-task learning for sequence labeling by incorporating a character-aware neural language model.
%incorporate a character-aware neural language model incorporates a neural language model and conducts multi-task learning to guide sequence labeling;
%
% incorporate character-aware neural language models into the model to extract knowledge from raw texts and empower the sequence labeling task.
Zhang \etal~\cite{zhang2018learning} propose a multi-channel model to learn the tag dependency via a combination of word-level Bi-LSTM and tag LSTM.
%based on the combination of word-level Bi-LSTM and tag LSTM for the learning of tag dependency.
%
Besides, there also exist several Bi-GRU based sequence labeling models, \eg~\cite{Yang2016Transfer}.
However, these RNN-based architectures are poor in modeling discrete context dependencies.
In contrary, our proposed model is based on the Bi-LSTM-CRF architecture with self-attention mechanism
to model the discrete position-aware dependencies
for addressing the sequence labeling problem.

\subsection{Attention Mechanism}

%\paratitle{Attention Mechanism}.

\paratitle{Self-Attention}.
%
%Self-attention is a special case of the attention mechanism to flexibly capture both successive and discrete dependencies over a given sequence.
%
Here, we mainly focus on reviewing self-attention based methods.
Self-attention is a special case of the attention mechanism to flexibly capture both successive and discrete dependencies over a given sequence.
Indeed, many studies have devoted to research on how to utilize self-attention mechanisms
to improve the performance of several NLP tasks 
%Most of existing works using self-attention mechanisms favor to several NLP tasks
% 
through aligning scores of different elements within a sequence,
%which has been widely used for many tasks, 
such as reading comprehension~\cite{Cheng2016Long}, textual entailment~\cite{lin2017structured}, sentiment analysis~\cite{lin2017structured}, machine translation~\cite{Vaswani2017Attention}, language understanding~\cite{Tao2017DiSAN} and semantic role labeling~\cite{Tan2018Deep}.
Cheng \etal~\cite{Cheng2016Long} extend the LSTM architecture with self-attention to enable \emph{adaptive} memory usage during recurrence, which favors to several NLP tasks, ranging from \emph{sentiment analysis} to \emph{natural language inference}.
%\eg \emph{language modeling}, \emph{sentiment analysis}, and \emph{natural language inference}.
%
Lin \etal~\cite{lin2017structured} introduce a sentence embedding model with self-attention, in which a $2$-dimensional matrix is utilized to represent the embedding and each row of the matrix attends on a different part of the sentence. The model is applied to \emph{author profiling}, \emph{sentiment analysis} and \emph{textual entailment}, and yields a significant performance gain over other methods.
%and applied them to $3$ different tasks including author profiling, sentiment analysis and textual entailment.
%
Vaswani \etal~\cite{Vaswani2017Attention} propose a RNN/CNN free self-attention network to construct a \emph{sequence-to-sequence} (\ie seq2seq) model and achieve the state-of-the-arts in the neural machine translation (NMT) task.
Shen \etal~\cite{Tao2017DiSAN} employ self-attention to encode sentences and achieve great inference quality on a wide range of NLP tasks.

However, the purposes of these studies are different from the current work and thus will not be discussed in detail.
The most related work is proposed by Tan \etal~\cite{Tan2018Deep},
where they propose a deep neural architecture with self-attention mechanism
for \emph{semantic role labeling} task and achieves the excellent performance,
which inspire us to follow this line to apply self-attention to sequence labeling tasks
for better learning the \emph{word}-level context features
and modeling the discrete dependencies over a given sequence.

\paratitle{Position based Attention}.
%Modeling Position Information within Attention}
%
Attention mechanism has strong ability to model dependencies among tokens, but it cannot effectively make full use of the position information of the sequence in its structure.
Vaswani \etal~\cite{Vaswani2017Attention} propose a transformer model solely based on attention mechanism that achieves excellent performance for Neural Machine Translation (NMT) tasks, and they also point out the problem of neglecting the position information within attention in the existing methods.
%A transformer model based on attention mechanism is proposed by Vaswani \etal~\cite{Vaswani2017Attention} for neural machine translation (NMT) task, which achieves excellent performance
%transformer model solely based on attention mechanism that achieves excellent performance for Neural Machine Translation (NMT) tasks, and they first points out the problem of the loss of position information within attention.
%
As such, they consider to inject position information using timing signal approach to encode absolute position, and then embed it into the representation of the input sequence in pre-processing progress with attentions.
%In order to address it, they inject position information by using timing signal approach to generate absolute position encoding, and adding it to the input sequence before being processed by attention.
%
Following the success of Transformer, several subsequent studies using the Transformer architecture with the same strategy are proposed~\cite{Tan2018Deep}.
%Some subsequent work that utilize Transformer architecture in their model also adopts the same strategy~\cite{Tan2018Deep}. 
Show \etal~\cite{Shaw2018Self} extend the self-attention mechanism to take into account the representations of the \emph{relative} distances among sequence elements, and yields the substantial improvements in NMT task.
%efficiently consider representations of the relative distances between sequence elements, and produces substantial improvements on the NMT task.
%
Similarly, Sperber \etal~\cite{sperber2018self} model the \emph{relative} position information by strictly limit the scope of self-attention within their neighboring representations, which favors to the long-sequence acoustic modeling.
Nevertheless, these approaches solely take account of the absolute or relative position information independent of sequence tokens while neglecting its interactions with their input presentations.
%restricting self-attention to neighboring representations, which favors the long-sequence acoustic modeling. However, these approaches considers the absolute or relative position information independent of the sequence tokens and neglects its interactions with the input representations. 
In contrast, our proposed \emph{position}-aware self-attention model explore the positional information of the given sequence in a more flexible manner, \ie mainly focusing on modeling of discrete context dependencies of that sequence.
%The proposed \emph{position}-aware self-attention in our model explores the positional information of an input sequence with a more flexible manner, focusing on the modeling of discrete context dependencies of sequence.

%%%%%%%%%%%%%%%%%%%%%%%%%%%%%%%%%%%%%%%%%%%%%%%%%%%%%%%
%%%%%%%%%%%%%%%%%   SECTION 2  %%%%%%%%%%%%%%%%%%%%%%%%
%%%%%%%%%%%%%%%%%%%%%%%%%%%%%%%%%%%%%%%%%%%%%%%%%%%%%%%

\section{Preliminary}
\label{sec:prelimi}

Typically, sequence labeling can be treated as a set of independent classification tasks,
which makes the optimal label for each member and then the global
best set of labels is chosen for the given sequence at once.
%select the globally best set of labels for 
%
Suppose we have a sequence ($\hat{x}$) composed of $n$ tokens, \ie $\hat{x}=[x_1, x_2, \ldots, x_n]^{\top}$,
we aim to assign a tag to each member
and output the corresponding globally best label sequence $\hat{y}=[y_1,y_2, \ldots, y_n]^{\top}$.
%
%However, accuracy is generally improved by making the optimal label for a given element dependent on the choices of nearby elements, using special algorithms to choose the globally best set of labels for the entire sequence at once.
%As an example of why finding the globally best label sequence might produce better results than labeling one item at a time, consider the part-of-speech tagging task just described. Frequently, many words are members of multiple parts of speech, and the correct label of such a word can often be deduced from the correct label of the word to the immediate left or right.
%
%the aim of sequence labeling is to assign a tag for each word and output the corresponding ground truth label sequence $\hat{y}=(y_1,y_2, \ldots, y_n)$.
%
Many neural models are proposed for
this task~\cite{Lample2016Neural,Ma2016End}.
By following the success of the state-of-the-art neural network architecture,
we briefly describe a \emph{Bi-LSTM-CRF} model for this task,
which often consists of three major stages:
%
%Following the recent state-of-the-art models for sequence labeling tasks ~\cite{Plank2016Multilingual,Lample2016Neural,Ma2016End}, we will use a Bi-directional LSTM-CRF model as our baseline. The model consists of mainly three stages:

\paratitle{Distributed Representation}, represents words in low dimensional real-valued dense vectors,
where each dimension represents a latent feature. Besides pre-trained word embeddings for the basic input, several studies also incorporate character-level representations for exploiting useful intra-word information (\eg prefix or suffix).
% such as prefix and suffix.

\paratitle{Context Encoder}, captures the context dependencies and learns contextual representations for tag decoding.
Traditional methods easily face the risk of gradient vanishing/exploding problem,
and thus several variants of RNNs, \eg LSTMs~\cite{hochreiter1997long},
are widely employed to be the context encoder architecture for different sequence labeling tasks,
owing to their promising performance on handling such problems.
%
%As promising performance on handling the gradient vanishing/exploding problem, variants (\ie LSTMs~\cite{hochreiter1997long}) of RNNs are widely employed to be the context encoder architecture for different sequence labeling tasks.
%
%gradient vanishing/exploding problem (handle/appear)
%%%
%LSTMs~\cite{Schuster1997Bidirectional} are variants of RNNs designed to cope with gradient vanishing/exploding problems and has become a widely-used context encoder architecture in sequence labeling tasks.
Therefore, here we briefly illustrate a special case of LSTM-CRF model, \ie \emph{Bi-directional} LSTM-CRF, which
incorporate past/future contexts from both directions (forward/backward) to generate the hidden states
of each word, and then jointly concatenate them to represent the \emph{global} information of the entire sequence.
%
%it is crucial to incorporate context from both directions
%
%generally generates forward/backward hidden states to model the past/future context information for each word, and then both two hidden states are concatenated to be as the \emph{global} information of the entire sequence.

However,
the sequential way to process the inputs of RNNs
might weaken the sensitivity of modeling discrete context dependencies,
since it recursively compose each word with its forward/backward hidden state that encodes the entire history/future information.
As such, the latent relationship between each pair of words is not well extracted, which is closely related to the final prediction task.
To this end, in this paper we propose a self-attentional context fusion layer
to better capture the relations among tokens and help to model discrete context dependencies,
via incorporating the complementary context information at different layers
in our proposed neural architecture. We will detail it in the following
sections, respectively.

\paratitle{Tag Decoder},
employs a CRF layer to produce a sequence of tags corresponding to the input sequence.
%To make the optimal label for each element of a given sequence,
Typically, the correct label to each element of a given sequence
often depends on the choices of nearby elements.
As such, the correlations between labels of adjacent neighborhoods are usually
considered for jointly decoding the best chain of labels for the entire sequence.
%the correct label of such member is often depended on the choices of nearby elements,
% and it is thus beneficial to consider the correlations between labels of adjacent neighborhoods, and then jointly decode the best chain of labels for the entire sequence.
%
%For a sequence labeling task, it is beneficial to consider the correlations between labels in neighborhoods and jointly decode the best chain of labels for a given sequence.
%
Additionally, CRF model has been proven~\cite{lafferty2001conditional} to be powerful
in learning the strong dependencies across output labels,
thus it is usually employed to make the optimal label for each element of the input sequence.
Specifically,
%Conditional random field (CRF)~\cite{lafferty2001conditional} has been included in most top-performing models to learn the strong dependencies across output labels and model label sequence jointly.
%
let $\matrix{Z} = [\hat{z}_{1},\hat{z}_{2}, \ldots,\hat{z}_{n}]^{\top}$
be the output of context encoder of the given
%training instance
sequence $\hat{x}$,
and thus the probability $\Pr(\hat{y}|\hat{x})$ of generating the whole label sequence $\hat{y}$ with regard to $\matrix{Z}$
is calculated by CRF model~\cite{Liu2017Empower},
%
%$(x_i,y_i)$, we suppose the output of context encoder is $\matrix{Z}_i = [\hat{z}_{i,1},\hat{z}_{i,2}, \ldots,\hat{z}_{i,n}]^{\top}$. CRF models describe the probability $\Pr(\hat{y}|x_i)$ of generating the whole label sequence $y_i\in\hat{y}$ with regard to $Z$, and the probability is defined as follows:
%
\begin{equation}
  \Pr(\hat{y}|\hat{x}) = \frac{\prod_{j=1}^{n}\phi(y_{j-1},y_j,\hat{z}_j)}
{\sum_{y^{'}\in{\matrix{Y}(\matrix{Z})}}\prod_{j=1}^n\phi(y^{'}_{j-1},y^{'}_j,\hat{z}_j)},
\end{equation}

\begin{equation}
	\phi(y_{j-1},y_j,\hat{z}_j)\!=\!\exp(\matrix{W}_{y_{j-1},y_{j}}\hat{z}_j + b_{y_{j-1},y_j}),
\end{equation}
where $\matrix{Y}(\matrix{Z})$ is the set of possible label sequences for $\matrix{Z}$;
$\matrix{W}_{y_{j-1},y_{j}}$ and  $b_{y_{j-1},y_j}$
indicate the weighted matrix and bias parameters corresponding to the label pair $(y_{j-1},y_j)$, respectively.
Then, we employ a likelihood function $\mathcal{L}$
to minimize the negative \emph{log} probability of the golden tag sequence for training,
%where the loss of the function is defined as $\mathcal{L}$:
%
\begin{equation}
\mathcal{L} = -\sum_{\hat{x}\in \mathcal{X};\hat{y}\in \mathcal{Y}}\log p(\hat{y}|\hat{x}),
\end{equation}
where $\mathcal{X}$ denotes the set of training instances, and $\mathcal{Y}$ indicates the corresponding
tag set.

\section{Proposed Approach}
\label{sec:method}

\begin{figure}[!t]
%\hspace{-0.4cm}
  \centering
	\includegraphics[width=0.5\textwidth]{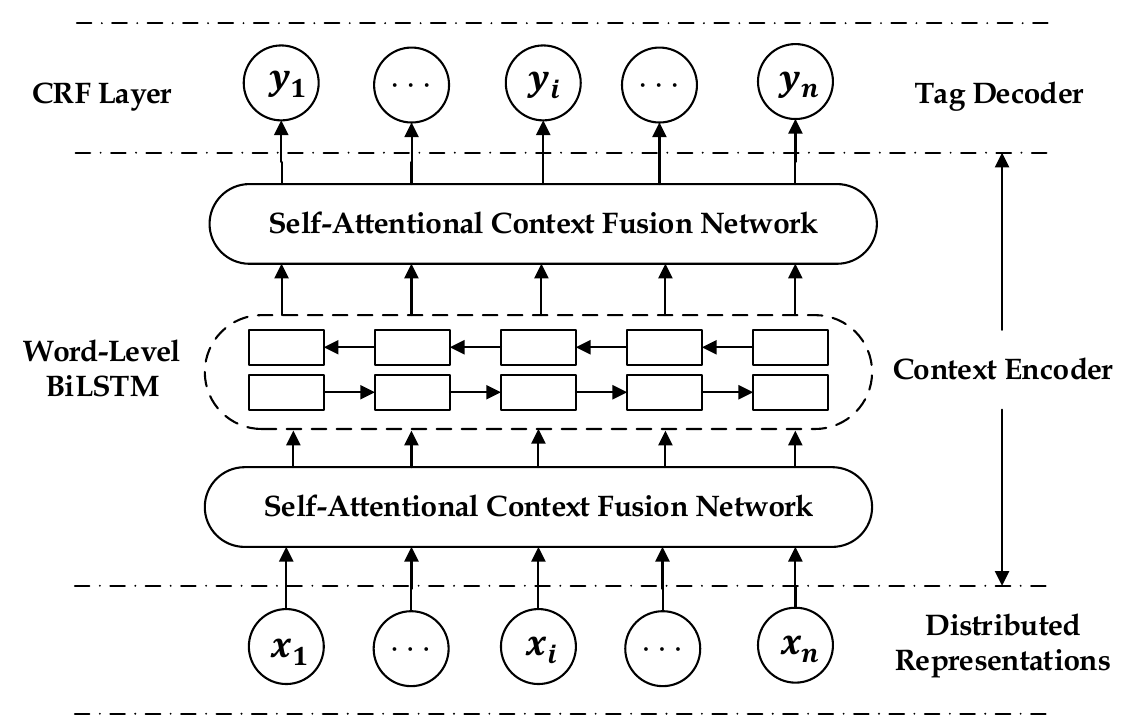}
	\caption{Overview of proposed neural architecture.}
	\label{fig:overview}
\end{figure}

%{\color{red}The neural architecture of our proposed model is visualized in Figure \ref{fig:overview}.}

As aforementioned,
RNN has limitations in modeling discrete context dependencies of the given sequence, thus
in this paper we mainly focus on how to effectively model this kind of context dependencies
%to provide the complementary context information
during the \emph{context encoder} stage
within LSTM-CRF architecture (\rf Section~\ref{sec:prelimi}).
%
%within the mentioned \emph{LSTM-CRF} architecture in Section~\ref{sec:prelimi}.
%
%More concretely,
Therefore, we propose a new neural architecture for sequence labeling (shown in Figure \ref{fig:overview}),
with a novel self-attentional context fusion layer that provides the complementary context information.
%
%To capture the relations and model context dependencies,
%
Specifically, there are two context fusion layers are incorporated
at different levels in our proposed architecture, \ie
the one is used for re-weighting the initial input (following the layer of distributed representations),
and the other is added for re-weighting the output of \emph{word}-level Bi-LSTM layer. The overall learning process of the proposed self-attentional context fusion network is illustrated in Algorithm 1.
%
%the proposed context fusion layers are incorporated at different levels in our proposed architecture, \ie one layer is added for re-weighting the initial input (following the layer of distributed representations), and the other is added for re-weighting the output of  \emph{word}-level Bi-LSTM layer.
%
Besides, a well-designed \emph{position}-aware self-attention mechanism with three different positional factors is also incorporated into the layer,
which models the discrete context dependencies via exploring the relative position information of tokens in a flexible manner. 
%with three different positional factors is incorporated into the layer, which is well designed to facilitate the modeling of discrete context dependencies by exploring relative position information of tokens in a flexible manner.

Next, we will elaborate our proposed sequence labeling model in detail.
More concretely, Section \ref{method:SA} will present the proposed \emph{position}-aware self-attention mechanism,
followed with the illustration of the proposed context fusion layer in Section \ref{method:SACFN}.
% we first illustrate the proposed \emph{position}-aware self-attention mechanism in Section \ref{method:SA}, and then present the proposed context fusion layer in Section \ref{method:SACFN}.

\begin{algorithm}
  \renewcommand{\algorithmicrequire}{\textbf{Input:}}
  \renewcommand{\algorithmicensure}{\textbf{Output:}}
  \caption{Learning Processes of Self-Attentional Context Fusion Network}
  \label{alg:1}
  \begin{algorithmic}[1]
    \REQUIRE The original token representations of sequence $\matrix{X}=[\hat{x}_1, \hat{x}_2, \ldots, \hat{x}_n]$
    \ENSURE The context-aware representations of sequence $\tilde{\matrix{X}}=[\tilde{\vec{x}}_1, \tilde{\vec{x}}_2, \ldots, \tilde{\vec{x}}_n]$

    \FOR{$i \in \{1 \ldots n\}$}
        \FOR{$j \in \{1 \ldots n\}$}
        \STATE \emph{// Compute the alignment score of $\hat{x}_i$ and $\hat{x}_j$} 
        \STATE Update $f(\hat{x}_{i},\hat{x}_{j})$ based on Equation \ref{eq:original_f} 
        \STATE \emph{// Compute the positional bias} 
        \STATE Update $\Psi_{ij}(\hat{x}_{i})$ based on Equation \ref{eq:7} 
        \STATE Update $f(\hat{x}_{i},\hat{x}_{j})$ by adding $\Psi_{ij}(\hat{x}_{i})$ 
        \ENDFOR
        \FOR{$j \in \{1 \ldots n\}$}
        \STATE  $a_i(j)$ = Softmax($f(\hat{x}_{i},\hat{x}_{j})$)
        \ENDFOR
        \STATE $\hat{s}_i \leftarrow Sum({a_i(j)\odot{\hat{x}_j}})$
        \STATE $\tilde{\vec{s}_i} \leftarrow Fullyconnect(\hat{s}_i)$
        \STATE \emph{// Fusion gate mechanism}
        \STATE Update $ \lambda$ based on Equation \ref{lambda} 
        \STATE Update $\tilde{\vec{x}_i}$ based on Equation \ref{final} 
    \ENDFOR
  \end{algorithmic}  
\end{algorithm}

\subsection{Position-Aware Self-Attention}
\label{method:SA}
In this section, we present a novel \emph{position}-aware self-attention
for better inducing the importance of each token to a specified token within the same sequence.
% which token is important to a specified token within the same sequence.
%

% 先声明3个bias 整合并加到f里面
Position modeling is benefit for optimizing the self-attention network,
since self-attention cannot encode position information of tokens in sequence.
%that position modeling is benefit for self-attention network due to the self-attention cannot encode position information of tokens in sequence.
%
Although the position information is implicitly encoded by LSTM in our neural architecture, the process of calculating alignment scores within self-attention is independent of the relative distance of tokens.
%However, the relative distance or position information is an important factor for inducing the latent relations among them. 
To this end, here
we explore the positional information of an input sequence to extend self-attention model with a different and novel method, 
aiming to better model the discrete context dependencies of sequence.
%aiming to favor the modeling of discrete context dependencies of sequence.
%
To be specific, we introduce three different \emph{positional} factors, \ie
\emph{self-disabled mask} bias $\matrix{M}_{ij}(.)$, \emph{distance-aware Gaussian} bias $\matrix{G}_{ij}(.)$ and \emph{token-specific position} bias
$\matrix{P}_{ij}(.)$, which are combined in a global positional bias function 
$\Psi_{ij}(.)$, and added to the baseline self-attention. The three factors are combined by
\begin{equation}
\label{eq:7}
  \Psi_{ij}(\hat{x}_{i})=\matrix{M}_{ij}(\hat{x}_{i})+\alpha\matrix{G}_{ij}(\hat{x}_{i})+(1-\alpha)\matrix{P}_{ij}(\hat{x}_{i}),
\end{equation}
where
$\alpha$ is a trainable trade-off parameter that controls the contributions of different biases.

%分别介绍三个bias术语的含义

The \emph{self-disabled mask} bias $\matrix{M}_{ij}(.)$ disables the attention of each token to itself, for better measuring its dependency on other tokens. The \emph{distance-aware Gaussian} bias $\matrix{G}_{ij}(.)$ considers the information of relative distance by utilizing the form of Gaussian distribution, and explicitly affect the computation of attention weights. 
The \emph{token-specific position} bias $\matrix{P}_{ij}(.)$ further addresses the interactions between the representations of relative positions and the input presentations, thus explores the relative distance in a more flexible manner.
The details of these three factors will be illustrated in the following sections.

More concretely, assume the token representations of sequence $\matrix{X}=[\hat{x}_1, \hat{x}_2, \ldots, \hat{x}_n]^{\top}$
with $\hat{x}_i\in \mathbb{R}^d$.
To measure the attention weight of each $\hat{x}_j$ to a specified token $\hat{x}_i$,
a compatibility function $f(\hat{x}_i,\hat{x}_j)$ is employed to
measure the pairwise similarity (\ie the alignment score)
of $\hat{x}_i$ and $\hat{x}_j$.

Many different self-attention mechanisms are proposed
but are different in the compatibility function $f(\hat{x}_i,\hat{x}_j)$,
here we adopt \emph{additive} attention mechanism~\cite{bahdanau2014neural},
which is implemented by a one-layer feed-forward neural network
and is often superior to others in practice, which is computed by
\begin{equation}
\label{eq:original_f}
f(\hat{x}_{i},\hat{x}_{j})=\hat{w}^{\top}\sigma(\matrix{W^{(1)}}\hat{x}_{i}+\matrix{W^{(2)}}\hat{x}_{j}+\hat{b}),
\end{equation}
where $\sigma(\cdot)$ is an activation function;
$\matrix{W^{(1)}},\matrix{W^{(2)}} \in \mathbb{R}^{d\times{d}}$ indict the weight matrices;
$\hat{w} \in \mathbb{R}^{d}$ is a weight vector, and $\hat{b}$ denotes
the bias vector.

For effectively encoding position information, we incorporate the proposed positional bias function 
$\Psi_{ij}(.)$ to $f(\hat{x}_{i},\hat{x}_{j})$, and the position-aware self-attention
is rewritten by
\begin{equation}
\label{eq:6}
f(\hat{x}_{i},\hat{x}_{j})=w^{\top}\sigma(\matrix{W^{(1)}}\hat{x}_{i}+\matrix{W^{(2)}}\hat{x}_{j}+\hat{b})+\Psi_{ij}(\hat{x}_{i}),
\end{equation}

Then the alignment score is converted by a \emph{softmax} function
with the normalization of all the $n$ elements within $\matrix{X}$, \ie
\begin{equation}
a_i(j)=\frac{\exp(f(\hat{x}_i,\hat{x}_j))}{\sum_{j^{'}}{\exp(f(\hat{x}_i,\hat{x}_{j^{'}}))}}.
\end{equation}

Finally, the output ($\hat{s}_i \in \mathbb{R}^d$) of the self-attention of $\hat{x}_i$ is a weighted sum of
representations of all tokens in $\matrix{X}$ according to the alignment scores, namely,
\begin{equation}
\label{eq:sum_attention}
\hat{s}_i=\sum_{j=1}^n{a_i(j)\odot{\hat{x}_j}}.
\end{equation}

\subsubsection{Self-Disabled Mask Bias}
For a specific token $x_i$, the goal of our self-attentional model is to measure its dependency on other tokens in the same sequence and further capture discrete context information,
thus it is benefit to prevent the interference of itself information when calculating alignment scores,
through disabling the attention of each token to itself.
As such, we adopt \emph{self-disabled mask}~\cite{Tao2017DiSAN}
for self-attention, which is
\begin{equation}
\matrix{M}_{ij}(\hat{x}_i)=
\begin{cases}
\ 0,& \text{$i\neq{j}$},\\
-\infty,& \text{$i=j$},
\end{cases}
\end{equation}
where $-\infty$ is used to neglect itself contribution in self-attention.
%our intention is to measure its dependency on other tokens, so disabling the attention of each token to itself is beneficial. Inspired by~\cite{Tao2017DiSAN}, we apply a diagonal-disabled mask such that:
%Here, adding $-\infty$ to the compatibility function before adopting softmax prevents each token from considering itself in the attention process.

\subsubsection{Distance-Aware Gaussian Bias}
\label{subsub:Gaussian}
Self-attention mechanism models the \emph{global} dependencies among input tokens regardless of their distance,
while the relative position information is important for modeling the \emph{local}
context in sequence labeling tasks.
Without loss of generality, we take \emph{POS tagging} as an example,
the POS tag of a word is more likely influenced by its neighbors,
as compared with other long-distance words.
In order to favor the modeling of short-range dependencies by self-attention, we
take account of a \emph{distance-aware Gaussian} bias to control the scope of local context of a specified token $x_i$,
and by incorporating it into the compatibility function, we make the relative distance among tokens to explicitly affect the computation of their corresponding attention weights. The distance-aware Gaussian bias is defined as
\begin{equation}
  \matrix{G}_{ij}(\hat{x}_{i})=\frac{-(i-j)^{2}}{2\varepsilon^{2}},
\end{equation}
where $i$, $j$ indicates the order of $\hat{x}_i$ and $\hat{x}_j$;
parameter $\varepsilon$ refers to the standard deviation that is empirically set as $\varepsilon=\frac{k}{2}$;
and $k$ is a window size, which is set as $10$ in our experiments.

\subsubsection{Token-specific Position Bias}
Gaussian bias only takes into account the information of relative distance among tokens,
however the way a relative distance affects the distribution of attention might not be the same for different focused tokens, and the discrete context dependencies within the sequence also have much diversity.
As such, modeling of the relative distance should be further explored in a more flexible manner for addressing the interactions between the representations of relative positions and the input presentations.
%further explanations of relative distance can be drawn in a more flexible manner by tackling interactions between the relative positions and input representations.
%

Inspired by Shaw's work~\cite{Shaw2018Self}, 
we learn a relative position representation matrix $\matrix{R} \in \mathbb{R}^{r\times{d}}$ and inject the position information into the attention score. Here $d$ denotes the representation dim and $r$ is a nonnegative value that reflects the maximum margin between two different tokens.
In other words,
the relative distance between two tokens would be clipped to $r$
if it is greater than the threshold,
following the essential hypothesis that the precise relative position information is not useful while beyond a certain distance.
And its value is equal to the window size $k$\footnote{Note that in the remainder it has the similar meaning when the context is clear and discriminative.}
Specifically,
%
%we apply a feed-forward network to transform $x_i$ into a hidden state,
%and then is mapped into a scalar $\matrix{P}_{ij}(x_{i})$ by conditioning on each token
a scalar $\matrix{P}_{ij}(\hat{x_{i}})$ is composed of two term, which are parameterized as follows,
%
% \begin{equation}
%   \matrix{P}_{ij}(\hat{x}_{i})=\hat{x}_{i}^{\top}\matrix{R}_{\mathcal{C}(i-j,r)}+g_{R}(\matrix{R}_{\mathcal{C}(i-j,r)}),
% \end{equation}
\begin{equation}
  \matrix{P}_{ij}(\hat{x}_{i})=
  \begin{cases}
  \ \hat{x}_{i}^{\top}\matrix{R}_{r}+(\hat{v}^{\top}\matrix{R}_r+\hat{b}),& \text{$\left|i-j\right|>r$},\\
    \ \hat{x}_{i}^{\top}\matrix{R}_{\left|i-j\right|}+(\hat{v}^{\top}\matrix{R}_{\left|i-j\right|}+\hat{b}),& \text{$\left|i-j \right| \le r$}.
  \end{cases}
\end{equation}
where $\hat{v} \in \mathbb{R}^{d}$ is a weight vector, and $\hat{b} \in \mathbb{R}^1$ denotes
the bias term.
Here the first term is computed by the inner product
of $\hat{x}_{i}$ and $\left|i-j\right|$-th (or $r$-th) element of $\matrix{R}$, which represent the content-dependent positional information. And the second term that transforms the corresponding representation of relative position to a scalar score, can be reparded as a general global position bias.
Note that the relative position representation matrix $\matrix{R}$ is also trainable which is optimized during training along with other parameters.

\subsection{Self-Attentional Context Fusion Layer}
\label{method:SACFN}
\begin{figure}[!t]
 \hspace{-3ex}
	\centering
	\includegraphics[width=0.5\textwidth]{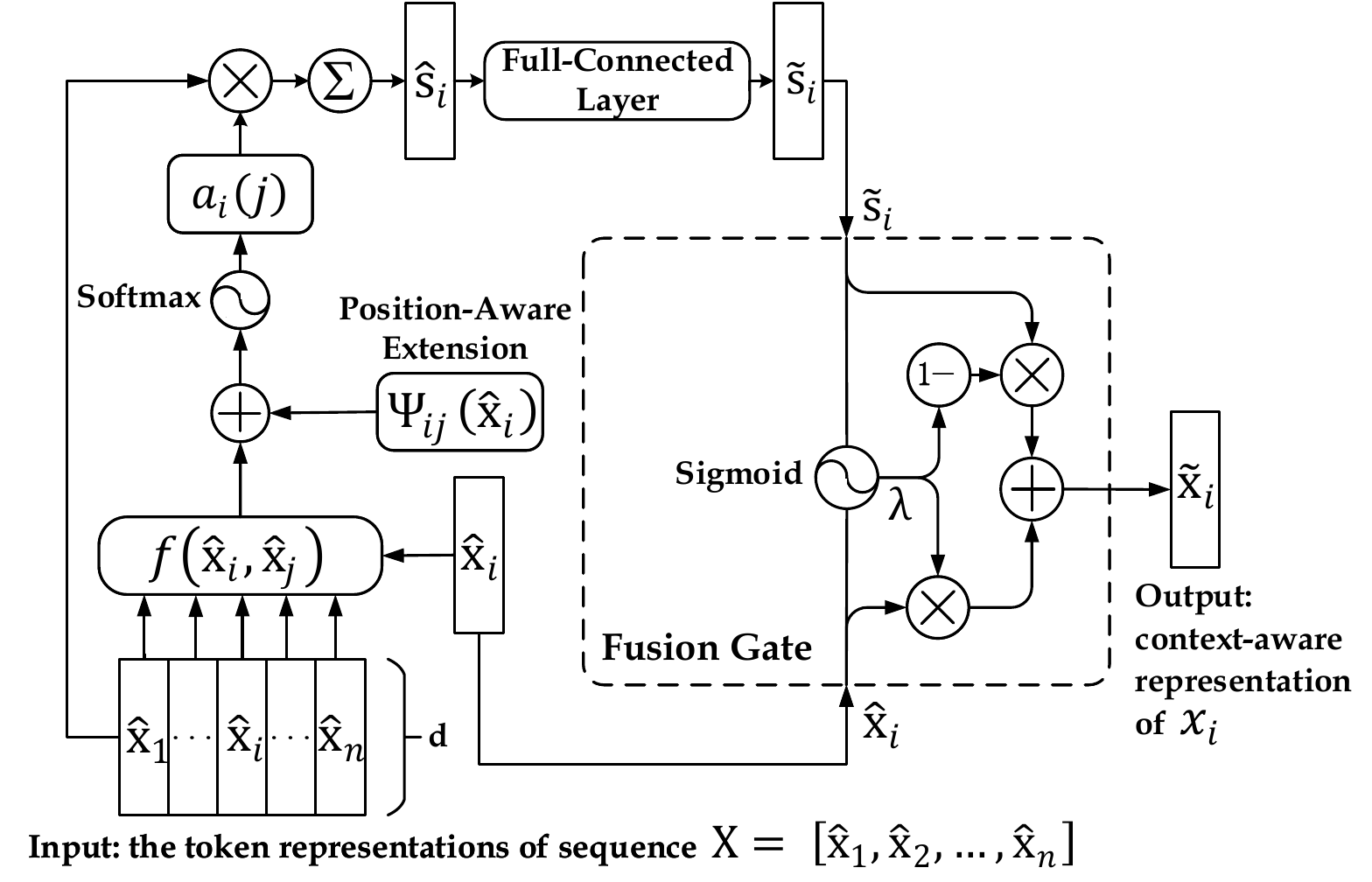}
	\caption{Self-Attentional Context Fusion Network.}
	\label{fig:SACFN}
 
\end{figure}

The success of neural networks stems from their highly flexible non-linear transformations.

Attention mechanism utilizes a weighted sum to generate the output vectors,
which limits its representational ability.
To further enhance the power of feature extraction of the attentional layer,
we take account of employing two fully connected layers to transform the outputs of the attention module,
which is formally computed by
%
%we have the following equation:
\begin{equation}
  \tilde{\vec{s}_i} = \tanh[\matrix{W^{(z2)}}\tanh(\matrix{W^{(z1)}}\hat{s}_i+\hat{b})],
\end{equation}
where $\matrix{W^{(z1)}}, \matrix{W^{(z2)}} \in \mathbb{R}^{d\times{d}}$ are trainable matrices;
and $\hat{s}_i$ denotes the output of the self-attention of $\hat{x}_i$ (\rf Eq \ref{eq:sum_attention}).

As we introduce a \emph{self-disabled mask} (\rf Section \ref{method:SA}) to disable the
attention of each token to itself,
the output of the proposed self-attention layer is insufficient for learning context-aware representation.
As such, we propose a feature-wise fusion gate mechanism to adaptively combine the feature of each token with its context.
Hence, the final context-aware representation (shown in Figure  \ref{fig:SACFN}) of $x_i$ is linearly combine with
input of the self-attention layer $\hat{x}_i$
and the output of the fully connected layers $\tilde{\vec{s}_i}$, namely
%
%a fusion gate to combine the information of each token with its context. Specifically, the fusion gate combines the input of the self-attention network $x_i$ and the output of the fully connected layers $\tilde{s_i}$ to produce final context-aware representations:
%
\begin{equation}
\label{lambda}
  \lambda = \mbox{sigmoid}(\matrix{W^{(f3)}}\tanh(\matrix{W^{(f1)}}\hat{x}_i+\matrix{W^{(f2)}}\tilde{\vec{s}_i}))
\end{equation}
\begin{equation}
\label{final}
  \tilde{\vec{x}_i} = \lambda\odot{\hat{x}_i}+(1-\lambda)\odot{\tilde{\vec{s}_i}}
\end{equation}
where $\matrix{W^{(f1)}}, \matrix{W^{(f2)}}, \matrix{W^{(f3)}} \in \mathbb{R}^{d\times{d}}$ are trainable weight matrices of the fusion gate. Note that the learned parameter $\lambda$ is a vector that has the same dimension with $\tilde{\vec{s}_i}$, because different features of $\tilde{\vec{s}_i}$ can contain different information of discrete context dependency. Hence the designed fusion gate is able to dynamically select useful information from the self-attention layer in a fine-grained manner.

\section{Experiments}
\label{sec:exp}

\subsection{Data Sets}

%\paratitle{Data Sets}.
%
We use four benchmark sequence labeling datasets for evaluation, \ie \emph{CoNLL 2003 NER dataset} (CoNLL03 NER)
, \emph{the Wall Street Journal portion of Penn Treebank dataset} (WSJ), \emph{CoNLL 2000 chunking dataset} (CoNLL00 chunking) and \emph{OntoNotes 5.0 English NER datasets} (OntoNotes 5.0).
The details about such corpora are shown in Table \ref{table1}.
%
%Here, we evaluate our model on two benchmark datasets: the CoNLL 2003 NER dataset~\cite{Sang2003Introduction} and the Wall Street Journal portion of Penn Treebank dataset (WSJ)~\cite{Treebank1993Building}. Table \ref{table1} list the data statistics of different corpus.
\begin{itemize}
\item \textbf{CoNLL03 NER} is a collection of news wire articles from the \emph{Reuters} corpus,
            which includes four different types of named entities: PER, LOC, ORG, and MISC.
            We use the standard dataset split~\cite{Collobert2011Natural} and follow \emph{BIOES} tagging scheme~(B, I, O, E, S).
            %It has been split into training, development and test sets, and following~\cite{Collobert2011Natural}, we adopt the BIOES tagging scheme~\cite{Ratinov2009Design}.
\item \textbf{WSJ} contains 25 sections and classifies each word into $45$ different types of POS tags. Here, we also adopt a standard data split method
            used in \cite{zhang2018learning}, namely, sections 0-18 as \emph{training} data, 19-21 as \emph{development} data, and sections 22-24 as \emph{test} data.
\item \textbf{CoNLL00 chunking} uses sections 15-18 from the Wall Street Journal corpus for \emph{training} and section 20 for \emph{testing}. It defines 11 syntactic chunk types (~\eg, NP, VP, ADJP) in addition to other. Following previous works~\cite{Peters2017Semi}, we randomly sampled 1000 sentences from the training set as \emph{development} data.

\item \textbf{OntoNotes 5.0} is much larger than \emph{CoNLL 2003 NER dataset}, and consists of text from a wide variety of sources (broadcast conversation, newswire, magazine, Web text, etc.). It is tagged with eighteen entity types (PERSON, ORG, GPE, LAW, etc.). Following previous works\cite{chiu2016named}, we adopt the portion of the dataset with gold-standard named entity annotations, in which the New Testaments portion is excluded. 

\end{itemize}

\setlength{\tabcolsep}{3pt}
\begin{table}[!t]
	\small
	\begin{center}
		\begin{tabular}{c|c|c|c|c}
			\hline
			\bf Corpus&\bf Type&\bf Train&\bf Dev&\bf Test\\
			\hline
			\hline
			{\multirow{2}{*}{\emph{CoNLL03 NER}}}&Sentences&14,987&3,466&3,684\\
			&Tokens&204,567&51,578&46,666\\
			
			\hline
			{\multirow{2}{*}{\emph{WSJ}}}&Sentences&38,219&5,527&5,462\\
			&Tokens&912,344&131,768&129,654\\
			\hline

			{\multirow{2}{*}{\emph{CoNLL00 chunking}}}&Sentences&8,936&1,000&2,012\\
			&Tokens&211,727&24,294&47,377\\
			\hline
			{\multirow{2}{*}{\emph{OntoNotes 5.0}}}&Sentences&59,924&8,528&8,262\\
			&Tokens&1,088,503&147,724&152,728\\
			\hline
		\end{tabular}
	\end{center}
	\caption{Statistics of \emph{CoNLL03 NER}, \emph{WSJ}, \emph{CoNLL00 chunking} and \emph{OntoNotes 5.0}.}
    \label{table1}
    \vspace{-0.2cm}
\end{table}

\subsection{Experimental Setting}

%\subsubsection{Model Settings}
%
We use LSTM to learn character-level representation of words,
and together with the pre-trained word embedding contributes to the distributed representation for input.
Then we initialize word embedding with $100$-dimensional GloVe~\cite{pennington2014glove} and randomly initialize $30$-dimensional character embedding.
Fine-tuning strategy is adopted
that we modify initial word embedding during gradient updates of the neural network model by back-propagating gradients.
%
%We randomly initialize word embedding with $100$-dimensional GloVe~\cite{pennington2014glove} and use $30$-dimensional character embedding which are randomly initialized.

The size of hidden state of character and word-level Bi-LSTM are set to $100$ and $300$, respectively. And we fix the number of Bi-LSTM layer as 1 in our neural architecture.
%
%The hidden state sizes of character/word-level Bi-LSTM are set to 100 and 300. Dropout strategy is adopted to mitigate overfitting on both the input and output vectors of Bi-LSTM with a rate of 0.55 and on output of self-attention with a rate of 0.2.
All weight matrices in our model are initialized by Glorot Initialization~\cite{glorot2010understanding}, and the bias parameters are initialized with $0$.
%The remaining of the initialization used in this paper are set according to \cite{Ma2016End}.
%We initialize the model parameters as in~\cite{Ma2016End}.

%\subsubsection{Optimization}
%
We train the model parameters by the \emph{mini-batch} stochastic gradient descent (SGD) with momentum.
The batch size and the momentum are set at 10 and 0.9, respectively.
The leaning rate is updated with $\eta_t = \frac{\eta_0}{1+\rho t}$, where $\eta_0$ is the initial learning rate ($0.01$ for POS tagging and $0.015$ for NER and chunking), $t$ is the number of epoch completed
and $\rho = 0.05$ is the decay rate.
The dropout strategy is used for overcoming the over-fitting on the input and the output
of Bi-LSTM with a rate of $0.55$, as well as the output of self-attention with a rate of $0.2$.
We also use gradient clipping~\cite{Pascanu2012On} to avoid gradient explosion problem. The threshold of gradient norm is set as $5.0$.
Early stopping~\cite{Lee2000Overfitting} is applied for training models according to their performances
on development sets.

\subsection{Evaluation Results and Analysis}
\subsubsection{Over Performance}
\begin{table}[!t]
	\small
	%\scriptsize
	\begin{center}
    \label{tabel2}
		\begin{tabular}{|c|c|c|}
			\hline
			{\multirow{2}{*}{\bf Index \& Model}}&\multicolumn{2}{c|}{\bf F1-score} \\
			\cline{2-3}&\bf Type&\bf Value($\pm $$\mbox{std}^{1}$)\\
			\hline
			\quad\quad Collobert et al., 2011~\cite{Collobert2011Natural}\quad\quad&reported&89.59\\
			\hline
		    \quad\quad Passos et al., 2014~\cite{Passos2014Lexicon}\quad\quad&reported&90.90\\
			\hline
		    \quad\quad Huang et al., 2015~\cite{Huang2015Bidirectional}&reported&90.10\\
			\hline
	        \quad\quad Lample et al., 2016~\cite{Lample2016Neural}\quad\quad&reported&90.94\\
			\hline
			\quad\quad Ma and Hovy, 2016~\cite{Ma2016End}\quad\quad&reported&91.21\\
			\hline
			\quad\quad Rei, 2017~\cite{Rei2017Semi}\quad\quad&reported&86.26\\
			\hline
			\quad\quad Zhang et al., 2017~\cite{Zhang2017Does}\quad\quad&reported&90.70\\
			\hline
		    \quad\quad Zhang et al., 2018~\cite{zhang2018learning}\quad\quad&reported&91.22\\
			\hline {\multirow{2}{*}{{{$\mbox{\quad\quad Liu et al., 2018\cite{Liu2017Empower}}^{2}$\quad\quad}}}}&avg&91.24$\pm$0.12\\
			\cline{2-3}&max&91.35\\
			\hline
			{\multirow{2}{*}{$\mbox{Bi-LSTM-CRF}^{3}$~\cite{Huang2015Bidirectional}}}&avg&91.01$\pm$0.21\\
			\cline{2-3}&max&91.27\\
			\hline
			{\multirow{2}{*}{Our model}}&avg&91.33$\pm$0.08\\
			\cline{2-3}&max&\textbf{91.42}\\
			\hline
			\hline
           \quad\quad Chiu and Nichols, 2016~\cite{chiu2016named}$^*$&reported&91.62$\pm$0.33\\
			\hline
			\quad\quad Yang et al., 2017~\cite{Yang2016Transfer}$^*$&reported&91.26\\
			\hline
            \quad\quad Peters et al., 2017~\cite{Peters2017Semi}$^*$&reported&91.93$\pm$0.19\\
			\hline
            \quad\quad Peters et al., 2018~\cite{Peters2018Deep}$^*$&reported&92.22$\pm$0.10\\
			\hline
			\quad\quad Devlin et al., 2018~\cite{devlin2018bert}$^*$&reported&92.80\\
			\hline
            \quad\quad Akbik et al., 2018~\cite{akbik2018contextual}$^*$&reported&93.09$\pm$0.12\\
			\hline
		\end{tabular}
	\end{center}
    \begin{tablenotes}
        \scriptsize
        \item[]{$^{1}~std$ means Standard Deviation.}
        \item[]{$^{2}$~Here we do not report the result used in~\cite{Liu2017Empower}, but update it with the result
        according to the first author's github~\url{https://github.com/LiyuanLucasLiu/LM-LSTM-CRF}, where the author claimed that the original
        result is not correct.}
        \item[]{$^{3}$~Here we re-implement the classical Bi-LSTM model using the same model setting and optimization method with our model.}
    \end{tablenotes}
    \caption{Comparison of overall performance on CoNLL 2003 NER task. Note that methods labelled with $\ast$ indicate that
    external knowledge are used and thus will not be compared for fairness in our experiments.}
    % utilized external knowledge beside CoNLL 2003 training set and pre-trained word embeddings.}}
    \label{table2}
\end{table}

\begin{table}[t!]
	\small
	\begin{center}
		\begin{tabular}{|c|c|c|}
			\hline
			{\multirow{2}{*}{\bf Index \& Model}}&\multicolumn{2}{c|}{\bf Accuracy} \\
			\cline{2-3}&\bf Type&\bf Value($\pm $std)\\
			\hline
			\quad\quad Collobert et al., 2011~\cite{Collobert2011Natural}&reported&97.29\\
			\hline
			\quad\quad Santos and Zadrozny, 2014~\cite{Santos2014Learning}&reported&97.32\\
			\hline
			\quad\quad Sun, 2014~\cite{sun2014structure}&reported&97.36\\
			\hline
            \quad\quad S{\o}gaard, 2011~\cite{S2011Semisupervised}&reported&97.50\\
			\hline
			\quad\quad Rei, 2017~\cite{Rei2017Semi}&reported&97.43\\	
			\hline
			\quad\quad Ma and Hovy, 2016~\cite{Ma2016End}&reported&97.55\\
			\hline
			\quad\quad Yasunaga et al., 2017~\cite{Yasunaga2018Robust}&reported&97.58\\
			\hline {\multirow{2}{*}{\quad\quad Liu et al., 2018~\cite{Liu2017Empower}}}&avg&97.53$\pm$0.03\\
			\cline{2-3}&max&97.59\\
			\hline
			\quad\quad Zhang et al., 2018~\cite{zhang2018learning}&reported&97.59\\
			\hline
			{\multirow{2}{*}{Bi-LSTM-CRF~\cite{Huang2015Bidirectional}}}&avg&97.51$\pm$0.04\\
			\cline{2-3}&max&97.56\\
			\hline
			{\multirow{2}{*}{Our model}}&avg&97.59$\pm$0.02\\
			\cline{2-3}&max&\textbf{97.63}\\
			\hline
		\end{tabular}
	\end{center}
%\vspace{-0.3cm}
	\caption{POS tagging accuracy of our model on test data from WSJ proportion of PTB, together with top-performance systems.}
    \label{table3}
     %\vspace{-0.2cm}
\end{table}

\begin{table}[!t]
	\small
	%\scriptsize
	\begin{center}
		\begin{tabular}{|c|c|c|}
			\hline
			{\multirow{2}{*}{\bf Index \& Model}}&\multicolumn{2}{c|}{\bf F1-score} \\
			\cline{2-3}&\bf Type&\bf Value($\pm$ std)\\
			\hline
			\quad\quad Collobert et al., 2011~\cite{Collobert2011Natural}\quad\quad &reported&94.32\\
			\hline
			\quad\quad Sun et al., 2014~\cite{sun2014feature}\quad\quad &reported&94.52\\
			\hline
            \quad\quad Huang et al., 2015~\cite{Huang2015Bidirectional}\quad\quad &reported&94.46\\
            \hline
            \quad\quad Ma and Sun, 2016~\cite{ma2016new}\quad\quad &reported&94.80\\
            \hline
			\quad\quad Rei, 2017~\cite{Rei2017Semi}\quad\quad &reported&93.88\\	
			\hline
            \quad\quad Zhai et al., 2017~\cite{zhai2017neural}\quad\quad &reported&94.72\\	
			\hline
            \quad\quad Zhang et al., 2017~\cite{Zhang2017Does}\quad\quad &reported&95.01\\	
			\hline	{\multirow{2}{*}{Bi-LSTM-CRF~\cite{Huang2015Bidirectional}}}&avg&94.92$\pm$0.08\\
			\cline{2-3}&max&95.01\\
			\hline
			{\multirow{2}{*}{Our model}}&avg&95.09$\pm$0.04\\
			\cline{2-3}&max&\textbf{95.15}\\
			\hline
			\hline
			\quad\quad Yang et al., 2017~\cite{Yang2016Transfer}$^*$\quad\quad &reported&95.41\\
			\hline
            \quad\quad Peters et al., 2017~\cite{Peters2017Semi}$^*$\quad\quad &reported&96.37$\pm$0.05\\
			\hline
            \quad\quad Akbik et al., 2018~\cite{akbik2018contextual}$^*$\quad\quad &reported&96.72$\pm$0.05\\
			\hline
		\end{tabular}
	\end{center}
    \caption{Comparison of overall performance on CoNLL00 chunking datatset. Note that methods labelled with $\ast$ indicate that
    external knowledge are used and thus will not be compared for fairness in our experiments.}
    % utilized external knowledge beside CoNLL 2003 training set and pre-trained word embeddings.}}
    \label{table_chunk}
\end{table}

\begin{table}[!t]
	\small
	%\scriptsize
	\begin{center}
		\begin{tabular}{|c|c|c|}
			\hline
			{\multirow{2}{*}{\bf Index \& Model}}&\multicolumn{2}{c|}{\bf F1-score} \\
			\cline{2-3}&\bf Type&\bf Value($\pm$ std)\\
			\hline
			\quad\quad Durrett and Klein, 2014~\cite{durrett2014joint}\quad\quad &reported&84.04\\
			\hline
			\quad\quad Chiu and Nichols, 2016~\cite{chiu2016named}\quad\quad &reported&86.28$\pm$0.26\\
			\hline
            \quad\quad Strubell et al., 2017~\cite{strubell2017fast}\quad\quad &reported&86.84$\pm$0.19\\
            \hline
            \quad\quad Li et al., 2017~\cite{li2017leveraging}\quad\quad &reported&87.21\\
            \hline
			\quad\quad Shen et al., 2018~\cite{shen2017deep}\quad\quad &reported&86.63$\pm$0.49\\	
			\hline
            \quad\quad Ghaddar and Langlais, 2018~\cite{ghaddar2018robust}\quad\quad &reported&87.95\\	
			\hline	{\multirow{2}{*}{Bi-LSTM-CRF~\cite{Huang2015Bidirectional}}}&avg&87.64$\pm$0.23\\
			\cline{2-3}&max&87.80\\
			\hline
			{\multirow{2}{*}{Our model}}&avg&88.12$\pm$0.22\\
			\cline{2-3}&max&\textbf{88.33}\\
			\hline
		\end{tabular}
	\end{center}
    \caption{Comparison of overall performance on OntoNotes 5.0 English NER datasets.}
    % utilized external knowledge beside CoNLL 2003 training set and pre-trained word embeddings.}}
    \label{table_ontonotes}
\end{table}

%\paratitle{Over Performance}.
%
This experiment is to evaluate the effectiveness of sequence labeling on different datasets by our approach.
Specifically, we report standard F1-score for CoNLL 2003 NER, CoNLL 2000 chunking and OntoNotes NER tasks,
and accuracy for POS tagging task on WSJ.
In order to enhance the fairness of the comparisons and verify the solidity of our improvement, we rerun $5$ times with different random initialization and report both average and max results of our proposed model as well as our re-implemented Bi-LSTM-CRF baseline.
The comparison methods used in this work are the state-of-arts in recent years that usually compared in many previous work.
The results for these four tasks are given in Table \ref{table2}, Table \ref{table3}, Table \ref{table_chunk} and Table \ref{table_ontonotes}, respectively.
Note we do not compare all of models listed in Table \ref{table2} and Table \ref{table_chunk},
as such methods (with $\ast$) utilize external knowledge excluding in the setting of training set,
% and pre-trained word embeddings,
like \emph{character type} and \emph{lexicon} features~\cite{chiu2016named},
shared information learned from other tasks~\cite{Yang2016Transfer},
other language models pre-trained from large unlabeled corpus~\cite{devlin2018bert}.
%which does not hold in the general setting of our proposed method.

Specifically, among the models listed in these tables,
Collobert et al.\cite{Collobert2011Natural} employ a simple feed-forward neural network with a fixed-size window for context feature extraction,
and adopt CRF method for jointly label decoding;
Huang et al.~\cite{Huang2015Bidirectional} introduce a Bi-LSTM-CRF model and outperform~\cite{Collobert2011Natural} by $0.51\%$ and $0.14\%$ on the dataset of CoNLL03 NER and CoNLL00 chunking, 
respectively, 
%since Bi-LSTM has a strong ability to extract long-range context features;
since Bi-LSTM has good characteristics in modeling sequential data and can better capture contextual information than window-based feed-forward neural network;
Lample et al.~\cite{Lample2016Neural} utilize the same architecture as baseline and further apply a LSTM layer to extract character level features of words, which outperform ~\cite{Huang2015Bidirectional} by $0.84\%$ for CoNLL03 NER task;
Similarly, Ma and Hovy~\cite{Ma2016End} achieve a significant improvement of $1.11\%$ over~\cite{Huang2015Bidirectional} on CoNLL03 NER dataset by equipping the Bi-LSTM-CRF model with a CNN layer to obtain character-level representations of words, which indicates the importance of exploiting useful intra-word information, and their proposed model also becomes a popular baseline for most subsequent work in this field;
Zhang et al.~\cite{Zhang2017Does} propose a method called Multi-Order BiLSTM which combines low order and high order LSTMs together in order to learn more tag dependencies, and this method outperforms~\cite{Huang2015Bidirectional} by $0.6\%$ and $0.55\%$ on the dataset of CoNLL03 NER and CoNLL00 chunking, however, it yields a worse performance than~\cite{Ma2016End};
Zhang et al.~\cite{zhang2018learning} propose a multi-channel model that performs better than~\cite{Ma2016End} with a slight improvement of $0.01\%$ and $0.04\%$ on CoNLL03 NER and WSJ dataset, which takes the long range tag dependencies into consideration by incorporating a tag LSTM in their model;
Liu et al.~\cite{Liu2017Empower} incorporate character-aware neural language models into the Bi-LSTM-CRF model and outperform ~\cite{Ma2016End} by $0.02\%$
on CoNLL03 NER task, but fail to achieve a better performance for POS tagging.

Note the results show that our proposed model outperforms Bi-LSTM-CRF model by $0.32\%$, $0.08\%$, $0.17\%$ and $0.48\%$ for the dataset of CoNLL03 NER, WSJ POS tagging, CoNLL00 chunking and OntoNotes 5.0, respectively, which could be viewed as significant improvements in the filed of sequence labeling. Even compared with the top-performance popular baseline~\cite{Ma2016End}, our model achieves a much better result for both NER and POS tagging tasks than other top-conference work in recent two years~\cite{zhang2018learning}, with an improvement of $0.12\%$ and $0.04\%$, respectively. Besides, the std (Standard Deviation) value of our model is smaller than the one of Bi-LSTM-CRF, which demonstrates our proposed method is more robust.
We also observe that our model consistently outperforms all these baselines for different tasks.
Because such models mostly adopt Bi-LSTM as their context encoder architecture, which cannot directly induce the
relations among two words, and thus omit modeling part of context dependency
especially some discrete patterns.
By proposing a novel \emph{position}-aware self-attention and incorporating self-attentional context fusion layers into the neural architecture, our proposed model is capable of
extracting the sufficient latent relationship among words, thus can provide the complementary context information on the basis of Bi-LSTM.
%the large improvements on the two datasets just verify the ability of our model to better model context dependencies.

%\section{Discussion}
%\label{sec:discuss}
%In this section we carry out a comparative analysis and wrap up the %discussion over the obtained results under a more qualitative view.

\subsubsection{Ablation Study}
In this section, we run experiments on the CoNLL 2003 NER dataset
to dissect the relative impact of each modeling decision by ablation studies.

%Without loss of generality, we task POS tagging task as an example,
For better understanding the effectiveness of our proposed \emph{position}-aware self-attention
in our model, we evaluate the performance of various position modeling strategies.
%
%our proposed method to model position information within self-attention
%Then we run experiments on \emph{CoNLL03 NER} dataset to dissect the effectiveness of each component in our proposed model by ablation study.
%To better understand the effectiveness of our proposed method to model position information within self-attention, we evaluate the performance of various position modeling strategies.
%
%
Training process is performed
$5$ times, and then the average F1-scores
%The mean F1-scores of 5 training repetitions
are reported in Table \ref{tabel_ablation1}.
Note that Model 3 is our final proposed architecture.
Model 1 remains the same as Model 3 except that it minus $\Psi_{ij}(\hat{x}_{i})$ in Eq \ref{eq:6}, which suggests there exists no position information within self-attention. Model 2 applies an absolute position encoding before context encoder layer on the basis of Model 1, which is the position modeling strategy adopted by Vaswani et al.~\cite{Vaswani2017Attention} in the Transformer model. Comparing Model 1 with Model 3, we can see that after removing the proposed positional bias $\Psi_{ij}(\hat{x}_{i})$ the performance decreases a lot, indicating that our proposed flexible extension of the self-attention achieves a significant improvement since it effectively explores the positional information of an input sequence. But Model 2 with absolute position encoding yields worse performance than Model 1. We conjecture that it is because the absolute position embedding might weaken model's ability to fusion context features in our architecture.
\begin{table}[t!]
	\small
	\begin{center}
		\begin{tabular}{ c|c|c }
			\hline
			\bf \quad No\quad & \bf \quad Model\quad & \bf \quad F1-score$\pm$std \quad \\
            \hline
			1&w/o $\Psi_{ij}(\hat{x}_{i})$ in Eq \ref{eq:6}&91.15$\pm$0.12\\
            \hline
			2&add position encoding&91.05$\pm$0.19\\
            \hline
            3&Our model&91.33$\pm$0.08\\
			\hline
		\end{tabular}
	\end{center}
%\vspace{-0.2cm}
	\caption{Experimental results of various position modeling strategies applied to self-attention.}
    \label{tabel_ablation1}
 %   \vspace{-0.3cm}
\end{table}

\begin{table}[!t]
	\small
	\begin{center}
		\begin{tabular}{ c|c|c|c }
			\hline
			\bf \quad $\matrix{M}_{ij}(\hat{x}_{i})$ \quad &\bf\quad $\matrix{P}_{ij}(\hat{x}_{i})$\quad & \bf\quad $\matrix{G}_{ij}(\hat{x}_{i})$ \quad& \bf\quad F1-score$\pm$std\quad\\
			\hline
			 $\times$&\checkmark&\checkmark&91.12$\pm$0.21\\
			\hline
            \checkmark& $\times$&\checkmark&91.07$\pm$0.05\\
			\hline
            \checkmark&\checkmark& $\times$&91.19$\pm$0.24\\
            \hline
            \checkmark&\checkmark&\checkmark&91.33$\pm$0.08\\
			\hline
		\end{tabular}
	\end{center}
     %\vspace{-0.3cm}
	\caption{Experimental results for ablating three positional factors.}
    \label{table_threebias}
    %\vspace{-0.3cm}
\end{table}

\begin{table}[!t]
	\small
	\begin{center}
		\begin{tabular}{ c|c|c }
			\hline
			\bf\quad First layer \quad& \bf \quad Second layer\quad & \bf\quad F1-score$\pm$std\quad \\
			\hline
            $\times$ &$\times$ &91.01$\pm$0.21\\
            \hline
			 $\times$&\checkmark&91.13$\pm$0.17\\
			\hline
            \checkmark& $\times$&91.27$\pm$0.05\\
            \hline
            \checkmark&\checkmark&91.33$\pm$0.08\\
			\hline
		\end{tabular}
	\end{center}
     %\vspace{-0.3cm}
	\caption{Experimental results for ablating two self-attentional context fusion layer.}
    \label{tabel_ablation2}
    %\vspace{-0.3cm}
\end{table}

\begin{table}[t!]
	\small
	\begin{center}
		\begin{tabular}{c|c|c|c|c|c}
			\hline
{\multirow{2}{*}{\bf No}}&\multicolumn{3}{c|}{\bf Context Encoder}
&{\multirow{2}{*}{\bf Decoder}}
&{\multirow{2}{*}{\bf F1-score$\pm$std}}
\\
\cline{2-4}&{\bf Bottom} &{\bf Middle}&{\bf Top}
& ~ \\
\hline
1 & SAN & SAN & SAN & CRF & 90.25$\pm$0.19\\
\hline
2 & Bi-LSTM & Bi-LSTM & SAN & CRF & 91.06$\pm$0.10\\
\hline
3 & SAN & Bi-LSTM & Bi-LSTM & CRF & 91.19$\pm$0.12\\
\hline
4 & SAN & Bi-LSTM & SAN & CRF & 91.33$\pm$0.08\\
\hline
5 & SAN & Bi-LSTM & SAN & Softmax & 88.79$\pm$0.26\\
\hline
		\end{tabular}
	\end{center}
	\caption{Experimental results for adjusting the architecture of the proposed model (SAN denotes the proposed self-attentional context fusion network).}
    \label{table_architecture}
\end{table}

\begin{table}[!t]
	\small
	\begin{center}
		\begin{tabular}{ c|c|c }
			\hline
			\bf\quad Num of layers \quad &  \bf\quad Position modeling strategy\quad   & \bf\quad F1-score$\pm$std\quad \\
			\hline
			1&Absolute position embedding&88.7$\pm$0.23\\
			\hline
			1& Proposed positional bias&90.69$\pm$0.21\\
			\hline
			2&Absolute position embedding&88.79$\pm$0.39\\
			\hline
			2&Proposed positional bias&90.7$\pm$0.15\\
			\hline
			3&Absolute position embedding&88.57$\pm$0.23\\
			\hline
			3&Proposed positional bias&90.6$\pm$0.12\\
			\hline
		\end{tabular}
	\end{center}
	%\vspace{-0.3cm}
	\caption{Experimental results of the Transformer model.}
	\label{table_transformer}
	%\vspace{-0.3cm}
\end{table}

In order to better understand the working mechanism of our proposed \emph{position}-aware self-attention, we further analysis the influence of three different positional factors incorporated in it. One of the three factors is removed from proposed positional bias function (Eq \ref{eq:7}) each time and the results are shown in Table \ref{table_threebias}. We can clearly see that the final proposed model including all three factors achieves the best performance and ablating any one bias contributes to a worse score. It demonstrates the effectiveness of our well designed positional bias to explore the relative position information of tokens from different perspectives. The result also shows that after removing $\matrix{P}_{ij}(\hat{x}_{i})$, the F1-score decreases the most, indicating the \emph{token-specific position} bias leads to a significantly better performance since it considers the relative positions in a more flexible manner by tacking the interactions with input representations and has advantages in modeling discrete context dependencies. 
% As for the trade-off parameter $\alpha$ in Eq \ref{eq:7}, we randomly select $20$ settings and finally find the model achieves best on the development set when $0.3$, $0.4$ and $0.3$ is assigned to $\alpha_1$, $\alpha_2$ and $\alpha_3$ respectively. 

In addition, in order to investigate the influence of our designed self-attentional context fusion layer, we also conduct ablation tests where one of the two layers (\rf Figure \ref{fig:overview}) is removed from our neural architecture each time.
Table \ref{tabel_ablation2} shows that including either one self-attentional context fusion layer contributes to an obvious improvement over the baseline model, which verifies the effectiveness of our proposed self-attentional context fusion layer to provide the complementary context information at different levels and then enhance the prediction.
%We also evaluate the impact of ablating each of the two self-attentional context fusion layer in Figure \ref{fig:overview}.

As can be seen from Figure \ref{fig:overview}, the context encoder of proposed neural architecture consists of three parts, \ie bottom, middle and top. And we conduct corresponding experiment to illustrate why Bi-LSTM is placed between two self-attentional layers. The results are given in Table \ref{table_architecture}. Note that Model 4 is our final proposed architecture, and comparing Model 1 with Model 4, we can see that after removing the Bi-LSTM layer the performance decreases a lot, indicating the powerful ability of Bi-LSTM to capture sequential long-term dependencies. By comparing Model 2 and Model 3 with Model 4, it can be found that replacing either self-attentional layer in the architecture with Bi-LSTM will not lead to an improvement in results, further illustrating the effectiveness and rationality of the designed architecture. Finally, we also shows the necessity of adopting CRF instead of softmax as the decoder by comparing Model 5 and Model 4.

The Transformer model~\cite{Vaswani2017Attention} which is based on self-attention mechanism has been proven to have strong capabilities for feature extraction. We evaluate the transformer with different numbers of layers on CoNLL03 NER task, and the result is given in Table \ref{table_transformer}. In the experiment we adopt the transformer as the context encoder architecture and remain the distributed representations and tag decoder part of our model. And we also evaluate the performance of various position modeling strategies on the Transformer architecture, in which the proposed positional bias is used to replace the absolute position encoding. Table \ref{table_transformer} shows that all these models yield a poor performance that even worse than most of our baselines. We conjecture that it's because the transformer model may be sensitive to the hyper-parameters for different sequence labeling tasks, since there are lots of hyper-parameters like dimension of keys/queries/values, dimension of attention model, dimension of inner-layer, number of heads and etc. As for the setting of this experiment, 
the parameters are set to $64$, $512$, $1024$ and $8$, respectively. 
 However, it's obvious that changing the position modeling strategy leads to an greate improvement to the results, which further demonstrate the effectiveness of our proposed method to explore position information for sequence labeling tasks.

\setlength{\textfloatsep}{10pt}
\begin{figure}[!t]
	\subfigure[CoNLL03 NER]{
		\begin{minipage}[b]{0.23\textwidth}
			\includegraphics[width=\textwidth] {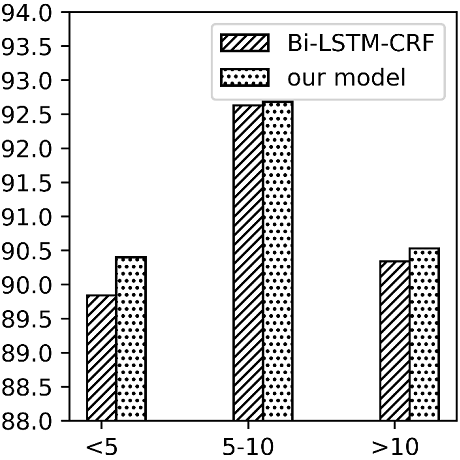}
		\end{minipage}
	}
	% \hspace{-0.9cm}
	\subfigure[WSJ POS tagging]{
		\begin{minipage}[b]{0.22\textwidth}
			\includegraphics[width=\textwidth]{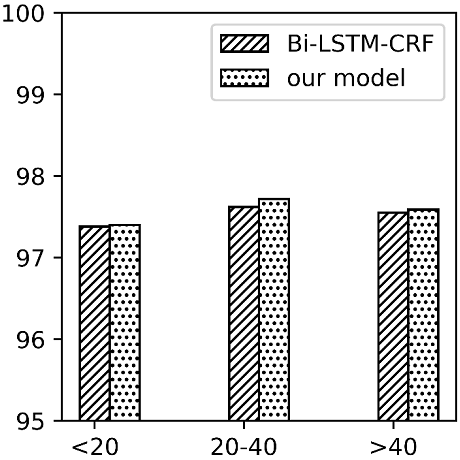}
		\end{minipage}
	}
	\subfigure[CoNL00 Chunking]{
		\begin{minipage}[b]{0.22\textwidth}
			\includegraphics[width=\textwidth]{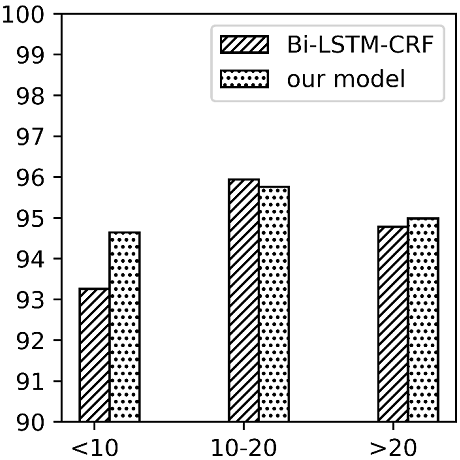}
		\end{minipage}
	}
	\subfigure[OntoNotes 5.0]{
		\begin{minipage}[b]{0.22\textwidth}
			\includegraphics[width=\textwidth]{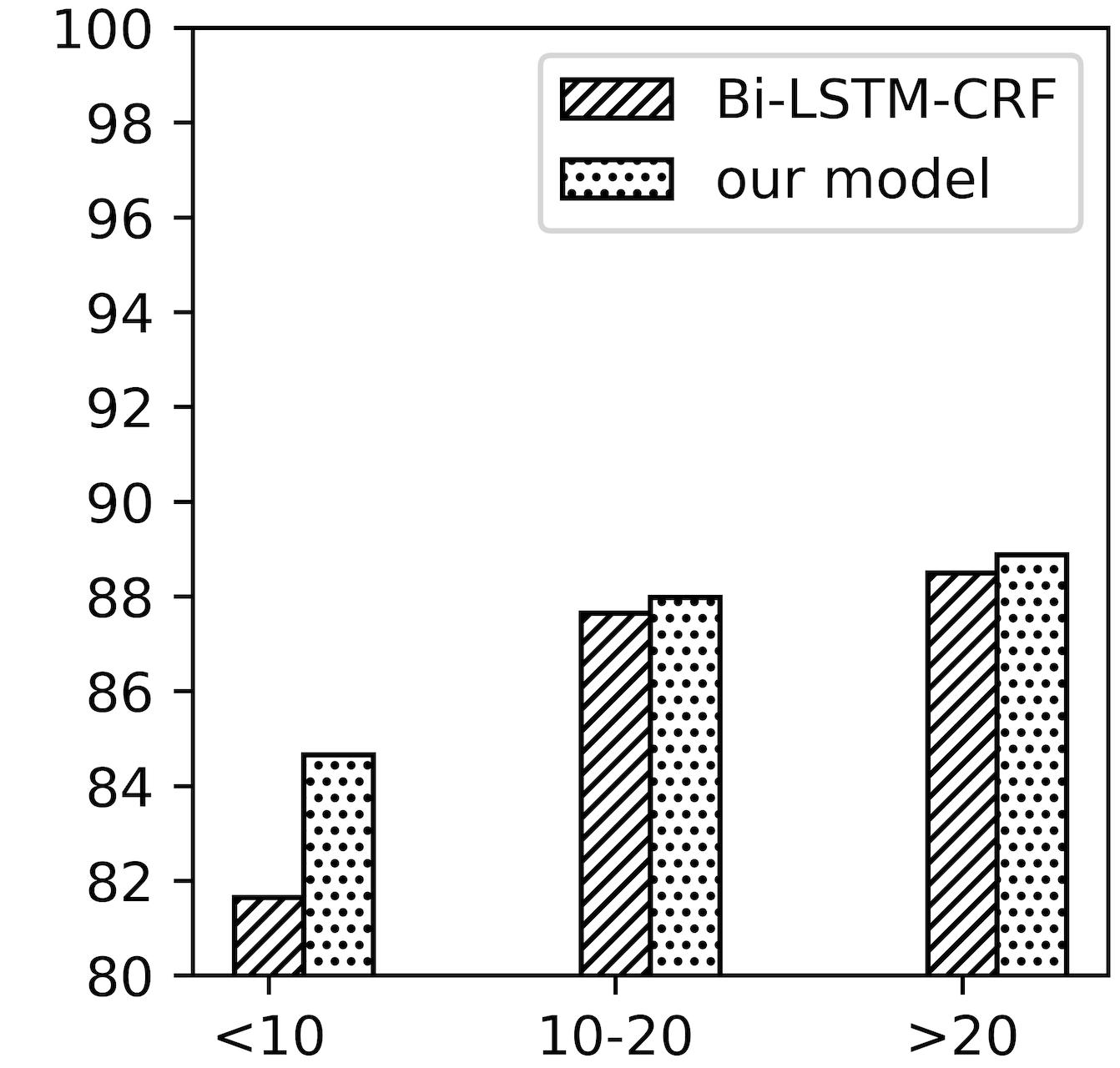}
		\end{minipage}
	}
	\caption{Performances on Different Lengths.}
	\label{fig:lengths}
    % \vspace{-cm}
\end{figure}

\subsubsection{Performances on Different Length}
We further analyze the performance of different models with respect to the different length of sentences. In Figure \ref{fig:lengths}, we compare Bi-LSTM-CRF baseline and our proposed model on different sentence lengths. For NER and chunking, our model significantly outperforms Bi-LSTM-CRF on short sentences (sentence length less than 5 on CoNLL2003, length less than 10 on CoNLL2000 and OntoNotes), which indicates that the improvement of the proposed model on short sentences is much larger than those on long sentences. The discrete context dependencies with short distances in a sequence are captured very well by our proposed model but simply neglected by Bi-LSTM-CRF. For POS tagging, performances of the two models on different sentence lengths are relatively comparable, while in the range of (20;40) our model performs slightly better than Bi-LSTM-CRF.

\begin{table*}
	\small
	\begin{center}
		\begin{tabular}{c|l}
			\hline
             Sent1 &The market opened sharply \textbf{lower}, with the Nikkei average down nearly 600 after 20 minutes. \\

             Gold &\,DT \quad NN \quad VBD \quad RB \quad  [RBR] \;IN \ \,DT \;NNP \quad NN \quad\  RB \quad RB \ \ \ CD \ \;IN \,CD \,\,NNS \\

             Bi-LSTM &\,DT \quad NN \quad VBD \quad RB \quad\;  \underline{JJR} \quad IN \ \,DT \;NNP \quad NN \quad\  RB \quad RB \ \ \ CD \ \;IN \,CD \,\,NNS \\

             Our Model &\,DT \quad NN \quad VBD \quad RB \quad [RBR] \;IN \ \,DT \;NNP \quad NN \quad\  RB \quad RB \ \ \ CD \ \;IN \,CD \,\,NNS \\
             \hline
             \hline
             Sent2 &The \; dollar  \; also  \; moved  \; \textbf{higher}  \; in Tokyo. \\

             Gold &\,DT \quad NN \quad RB \quad VBD \quad [RBR] \; IN \, NNP \\

             Bi-LSTM &\,DT \quad NN \quad RB \quad VBD \quad \; \underline{JJR} \quad \, IN \, NNP \\

             Our Model &\,DT \quad NN \quad RB \quad VBD \quad [RBR] \; IN \, NNP \\
             \hline
             \hline
             Sent3 &But the rally was confined to the stocks , which \,had \,\,been \,hard \ \textbf{hit} during Friday 's selling frenzy. \\
             Gold &\,CC DT \,NN \,VBD \;VBN \ TO DT \ NNS \ WDT VBD VBN \,RB [VBN] \,IN \ \ \ NNP \;POS NN \ \ \; NN \\
             Bi-LSTM &\,CC DT \,NN \,VBD \;VBN \ TO DT \ NNS \ WDT VBD VBN \,RB \;\;\underline{NN} \quad IN \ \ \ NNP \;POS NN \ \ \; NN \\
             Our Model &\,CC DT \,NN \,VBD \;VBN \ TO DT \ NNS \ WDT VBD VBN \,RB [VBN] \,IN \ \ \ NNP \;POS NN \ \ \; NN \\
            \hline
            \hline
            Sent4 &Spending patterns in newspapers have \; been \; \textbf{upset} \, by shifts \; in ownership and general hardships.
            \\
            Gold &\quad NN \quad \; NNS \quad IN \quad NNS \quad  VBP \;VBN \, [VBN] \;IN \; NNS   IN \quad NN \quad\; CC \quad \;JJ \quad \,NNS
            \\
            Bi-LSTM &\quad NN \quad \; NNS \quad IN \quad NNS \quad  VBP \;VBN \, \, \quad \underline{JJ} \;\; IN \; NNS   IN \quad NN \quad\; CC \quad \;JJ \quad \,NNS
            \\
            Our Model &\quad NN \quad \; NNS \quad IN \quad NNS \quad  VBP \;VBN \, [VBN] \;IN \; NNS   IN \quad NN \quad\; CC \quad \;JJ \quad \,NNS
            \\
            \hline
		\end{tabular}
	\end{center}
	\caption{Examples of the predictions of Bi-LSTM-CRF baseline and our model.}
    \label{table_case}
\end{table*}

\subsubsection{Impact of Window Size}

\begin{figure}[!t]
	\centering
	\includegraphics[width=0.5\textwidth]{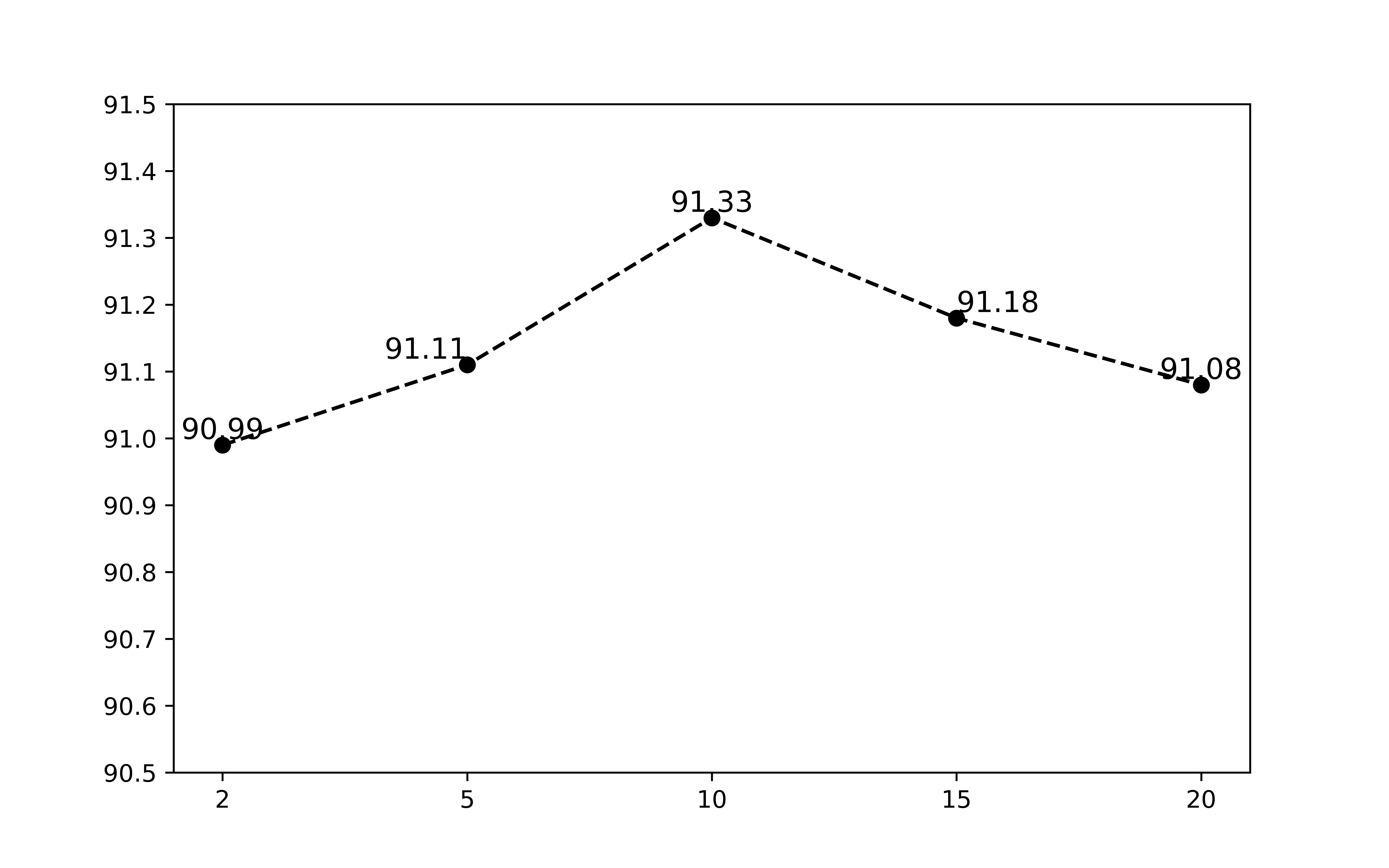}
	\caption{Performance of our model with various window sizes.}
	\label{fig:linechart}
\end{figure}

The window size $k$ (\rf Section \ref{subsub:Gaussian}) is clearly a hyperparameter which must be optimized for, thus we investigate the influence of the value of $k$ on the CoNLL2003 NER task. We also rerun $5$ times with different random initialization and report the average score, which is consistent with our other experiments in this paper. The plot in Figure \ref{fig:linechart} shows that when assigning the value of $k$ to $10$ we do outperform other models substantially. And with other window sizes (except $2$) our model performs relatively well and is superior to the Bi-LSTM-CRF baseline ($91.01\%$), which also suggests the effectiveness of our proposed \emph{distance-aware Gaussian} bias to favor the local context dependencies of sequence.

\subsubsection{Efficiency}

\begin{table}[!t]
	\small
	\begin{center}
		\begin{tabular}{ c|c|c|c }
			\hline
			\bf\quad Model \quad &\bf\quad F1-score$\pm$std &\bf\quad speed \quad & \bf\quad time \quad \\
			\hline
			Bi-LSTM-CRF&91.01$\pm 0.21$& $23$ iter/s &$1.6$ h\\
			\hline
			Out model&91.33$\pm 0.08$& $20$ iter/s &$1.9$ h\\
			\hline
		\end{tabular}
	\end{center}
	%\vspace{-0.3cm}
	\caption{Training speed, training time and perfromance of Bi-LSTM-CRF baseline and our proposed model on CoNLL 2003 NER task. $N$ iter/s means processing $N$ iterations per second.}
	\label{table_efficiency}
	%\vspace{-0.3cm}
\end{table}

\begin{table}[!t]
	\small
	\begin{center}
		\begin{tabular}{c|l}
			
			\hline
             Sent1 &Results of \quad Asian \quad Cup group C matches played on Friday \\

             Tag & \quad O \; \; O \; B-MISC  E-MISC O  \; O \quad O  \quad \; \:O \;\; \; \;O \;O \\

             $\lambda_1$ & \; 0.89 \;0.8 \; 0.61 \;\;\;\; 0.88 \; 0.75 0.86 0.87\; 0.87 \; 0.85 0.81\\

             $\lambda_2$ & \; 0.93 \;0.93\; 0.92\;\;\;\; 0.86 \; 0.87 0.94 0.89\; 0.92 \; 0.94 0.94 \\
             \hline
             \hline
			Sent2 &Japan \; - \; Hassan Abbas \;84 \;own\; goal\; , \;Takuya\; Takagi \;88\\

             Tag & S-LOC O B-PER E-PER \;O \; \;O \;\;\; O\;\; \;O \;B-PER \;E-PER \;O \\
			 $\lambda_1$ & \; 0.81 \;0.78 \; 0.73\; 0.7 \ \;0.89\; 0.83 \;0.89 \;0.77\; 0.74\; 0.72 \;0.89\\
             $\lambda_2$ & \; 0.93 \;0.87 \; 0.91\; 0.78\ \;0.83\; 0.93 \;0.93 \;0.93\; 0.9\; 0.77 \; 0.8 \\
             \hline
             \hline
             Sent3 &2. \; Candice \; Gilg  \; ( \; France ) 24.31 \\

             Tag & O \; B-PER \; E-PER O S-LOC O \; O \\

             $\lambda_1$ & 0.77 \; 0.74 \; 0.74 \; 0.75 0.79 0.73 0.76\\

             $\lambda_2$ & 0.89 \; 0.82\;  0.65 \; 0.56 0.77 0.71 0.87\\
             \hline
		\end{tabular}
	\end{center}
	\caption{Qualitative analysis of learned parameter $\lambda$ (\rf Eq \ref{final}). $\lambda_1$ and $\lambda_2$ denote $\lambda$ in the first and second self-attentional context fusion layer, respectively. Since $\lambda$ is a multi-dimensional vector, here we take its average in various dimensions to facilitate the observation.}
    \label{table_qua}
\end{table}

We implement our model based on the PyTorch library.
Models have been trained on one GeForce GTX 1080 GPU, with training time recorded in Table \ref{table_efficiency}.
In terms of efficiency, our model only introduces a small number of
parameters in two self-attention layer, which may not have a very large impact on efficiency.
And it can be drawn from Table \ref{table_efficiency} that the training speed of our model is
only $13\%$ lower than the baseline, but bring a significant improvement in the performance.

\subsubsection{Qualitative Analysis}

\setlength{\textfloatsep}{10pt}
\begin{figure}[!t]
\subfigure[Attention probability of word \emph{lower} for the first case]{
\begin{minipage}[b]{0.45\textwidth}
\includegraphics[width=\textwidth] {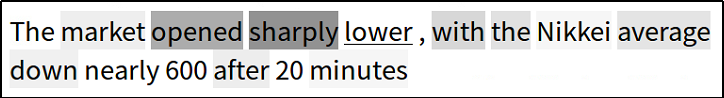}
\end{minipage}
\label{fig:case1}
}
\subfigure[Attention probability of word \emph{higher} for the second case]{
\begin{minipage}[b]{0.45\textwidth}
\includegraphics[width=\textwidth]{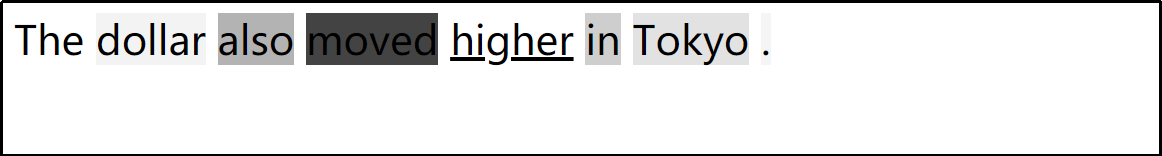}
\end{minipage}
\label{fig:case2}
}

\subfigure[Attention probability of word \emph{hit} for the third case]{
	\begin{minipage}[b]{0.45\textwidth}
		\includegraphics[width=\textwidth]{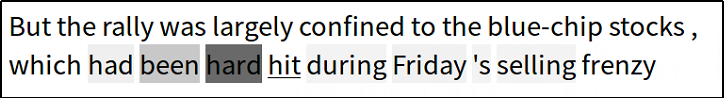}
	\end{minipage}
	\label{fig:case3}
}
\subfigure[Attention probability of word \emph{upset} for the fourth case]{
	\begin{minipage}[b]{0.45\textwidth}
		\includegraphics[width=\textwidth]{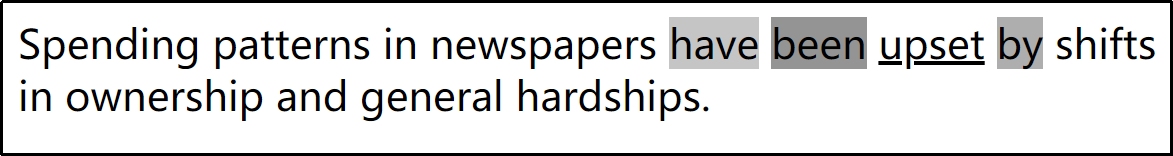}
	\end{minipage}
	\label{fig:case4}
}
\caption{Heatmaps of four cases.}
\end{figure}

The weight $\lambda$ in fusion gate mechanism actually indicates the balance between the feature of each token and its context representation obtained by self-attention.
If $\lambda$ is bigger than $0.5$, the final contextual representation relies more on its own feature, otherwise, the context representation play a more important role.
As shown in Table \ref{table_qua}, $\lambda$ varies along the sentence, showing the effectiveness of both feature.
Besides, we can observe that for most tokens, $\lambda_1$ is smaller than $\lambda_2$, which indicates that the first self-attentional layer before Bi-LSTM incorporates more useful contextual information.

\subsubsection{Case Study}

In this section, we present an in-depth analysis of results given by our proposed approach for better
understanding the influence of self-attention mechanism in our proposed model.
Without loss of generality, we take \emph{POS tagging} as the task and Bi-LSTM-CRF as the comparison method
for comparison.
Table \ref{table_case} shows four cases that our model predicts correctly but Bi-LSTM-CRF doesn't.
For better comparison, we visualize the alignment score by \emph{heat-maps} of words that baseline model fails to predict their labels correctly.

%In order to gain a closer view of how self-attention mechanism incorporated in our model influence the tagging result, we visualize the alignment score by heatmaps of words that baseline model fails to predict their labels correctly.

%Take part-of-speech tagging task for example,
%Table \ref{table_case} shows two cases that our model predicts correctly but baseline model doesn't.

%In Figure \ref{fig:case1},
In the first case,
the POS tag of ``lower" should be tagged with \emph{adverb comparative} (RBR),
while Bi-LSTM-CRF recognizes it as \emph{adjective comparative} (JJR).
It's obvious that the tag of ``lower" is dependent on the 3rd word ``open",
where an adverb is associated with a verb, and the 4th word ``sharply" is a direct modifier of it.
Figure \ref{fig:case1} shows that for word ``lower" it pays more attention on ``opened" and ``sharply",
while less on other words.
%get little attention from it.
Similar situation is shown in the second case, where our model assigns correct POS tag to ``higher" which depends largely on its previous word ``moved" but Bi-LSTM-CRF fails.
Regarding the third case,  our model succeeds in assigning \emph{verb past participle} (VBN) to word ``hit" by considering ``been" and ``hard" while Bi-LSTM-CRF makes a wrong decision.
The consistent conclusion is also reflected in Figure \ref{fig:case3}, that
``been" and ``hard" obtain large attention from the focus word ``hit".
And our model predict the POS tag of ``upset" correctly in the fourth case
which can be speculated from the common phrase ``have been done by".

Our analysis suggests if the choice of assigning label to a specified token $x_i$ depends on several other words,
they will receive a large amount of attention scores from $x_i$,
which also provides a high level interpretability for our self-attentional model.

\section{Conclusions}
\label{sec:conclusion}

This paper proposes a innovative neural architecture for sequence labeling tasks, in which a self-attentional context fusion layer is designed and incorporated to better model discrete and discontinuous context patterns of sequence.
%优点在于
The strengths of our work are that we identify the problem of modeling discrete context dependencies in sequence labeling tasks,
and a position-aware self-attention is proposed to
induce the latent independent relations among tokens over the input
sequence via three different
bias,
which can effectively model the context dependencies of
given sequence according to the relative
distance among tokens.
Experimental results on \emph{part-of-speech} (\textbf{POS}) \emph{tagging}, \emph{named entity recognition} (\textbf{NER}) and \emph{phrase chunking} tasks demonstrate the effectiveness of our proposed model which achieves \emph{state-of-the-art} performance. Furthermore, our analysis reveals the effects of each modeling decision from different perspectives. 
The way we model the discrete context dependencies of sentences in sequence labeling tasks can also inspire other researchers in the field to innovate from this perspective.
Despite the good performance, our work still has weaknesses, which is reflected in the limited improvement of our model for longer sequences. The main reason is that the second positional bias that we introduced tends to let self-attention learn the influence of neighboring words in the sequence.
In the future, we plan to further apply our neural architecture to data from other domains such as social media and empower more sequence labeling tasks. Additionally,
we also plan to employ our model to other sequence learning tasks besides sequence labeling, such as event extraction and neural machine translation. 
More recently, pretrained language models from huge corpus are widely adopted to enhance the representation of words. We will in the future explore integrating language modeling into this architecture to further boosting performance.

%\bibliography{main}

\bibliographystyle{IEEEtran}
\bibliography{main}

% \end{spacing}
\end{document}